\documentclass[lettersize,journal]{IEEEtran}
\usepackage{amsmath,amsfonts, amssymb}
\usepackage{pifont}
\newcommand{\cmark}{\ding{51}}%
\newcommand{\xmark}{\ding{55}}%
\usepackage{algorithmic}
\usepackage{algorithm}
\usepackage{array}
\usepackage[caption=false,font=normalsize,labelfont=sf,textfont=sf]{subfig}
\usepackage{textcomp}
\usepackage{stfloats}
\usepackage{url}
\usepackage{verbatim}
\usepackage{graphicx}
\usepackage{cite}
\usepackage{times}
\usepackage{epsfig}
\usepackage[accsupp]{axessibility}  
\usepackage{multirow}
\usepackage[table, dvipsnames]{xcolor}

\usepackage{booktabs}
\usepackage{soul}
\usepackage{cuted}
\usepackage{caption}
\usepackage{tabularx}
\usepackage{hyperref}
\usepackage{xspace}

\DeclareMathOperator*{\argmin}{arg\,min} 

\makeatletter
\DeclareRobustCommand\onedot{\futurelet\@let@token\@onedot}
\def\@onedot{\ifx\@let@token.\else.\null\fi\xspace}

\def\eg{\emph{e.g}\onedot} 
\def\ie{\emph{i.e}\onedot} 
\def\cf{\emph{c.f}\onedot} 
 
\def\wrt{w.r.t\onedot} 
\def\etal{\emph{et al}\onedot}
\makeatother

\begin{document}
\title{Privacy Preserving Federated Learning with Convolutional Variational Bottlenecks}

\author{Daniel~Scheliga,
        Patrick~M\"{a}der,
        and~Marco~Seeland
\thanks{This work was supported by the Thuringian Ministry of Economics, Science and Digital Society under Grant 5575/10-3.}
\thanks{The authors are with the Department of Computer Science and Automation, Technische Universit\"{a}t Ilmenau, 98693 Ilmenau, Germany (e-mail: daniel.scheliga@tu-ilmenau.de, patrick.maeder@tu-ilmenau.de, marco.seeland@tu-ilmenau.de}
}

\maketitle
\begin{abstract}
Gradient inversion attacks are an ubiquitous threat in federated learning as they exploit gradient leakage to reconstruct supposedly private training data.
Recent work has proposed to prevent gradient leakage without loss of model utility by incorporating a \textit{PRivacy EnhanCing mODulE} (PRECODE) based on variational modeling.
Without further analysis, it was shown that PRECODE successfully protects against gradient inversion attacks.
In this paper, we make multiple contributions.
First, we investigate the effect of PRECODE on gradient inversion attacks to reveal its underlying working principle.
We show that variational modeling introduces stochasticity into the gradients of PRECODE and the subsequent layers in a neural network.
The stochastic gradients of these layers prevent iterative gradient inversion attacks from converging.
Second, we formulate an attack that disables the privacy preserving effect of PRECODE by purposefully omitting stochastic gradients during attack optimization.
To preserve the privacy preserving effect of PRECODE, our analysis reveals that variational modeling must be placed early in the network. 
However, early placement of PRECODE is typically not feasible due to reduced model utility and the exploding number of additional model parameters.
Therefore, as a third contribution, we propose a novel privacy module -- the Convolutional Variational Bottleneck (CVB) -- that can be placed early in a neural network without suffering from these drawbacks.
We conduct an extensive empirical study on three seminal model architectures and six image classification datasets.
We find that all architectures are susceptible to gradient leakage attacks, which can be prevented by our proposed CVB.
Compared to PRECODE, we show that our novel privacy module requires fewer trainable parameters, and thus computational and communication costs, to effectively preserve privacy.
\end{abstract}


\section{Introduction}\label{sec:intro}
\IEEEPARstart{F}{ederated} Learning uses distributed data to collaboratively improve the utility of neural networks.
Multiple clients exchange local training gradients to collaboratively train a common global model.
This eliminates the need for centrally aggregated or shared data.
Because training data remains local to each participating client, such collaborative learning systems aim to systematically mitigate privacy risks~\cite{mcmahan2017communication, kairouz2021advances}.

However, recent studies show that potentially sensitive information can be reconstructed from the exchanged gradient information, thereby compromising the privacy of the clients.
Particularly advanced in this regard are iterative \textit{gradient inversion attacks}~\cite{zhu2019deep, zhao2020idlg, wei2020framework, geiping2020inverting, yin2021see}.
Such attacks are based on optimization of initially random dummy data to minimize a distance function between dummy gradients and the attacked client gradients.

\begin{figure}[!t]
	\begin{center}
		\includegraphics*[width=.99\linewidth]{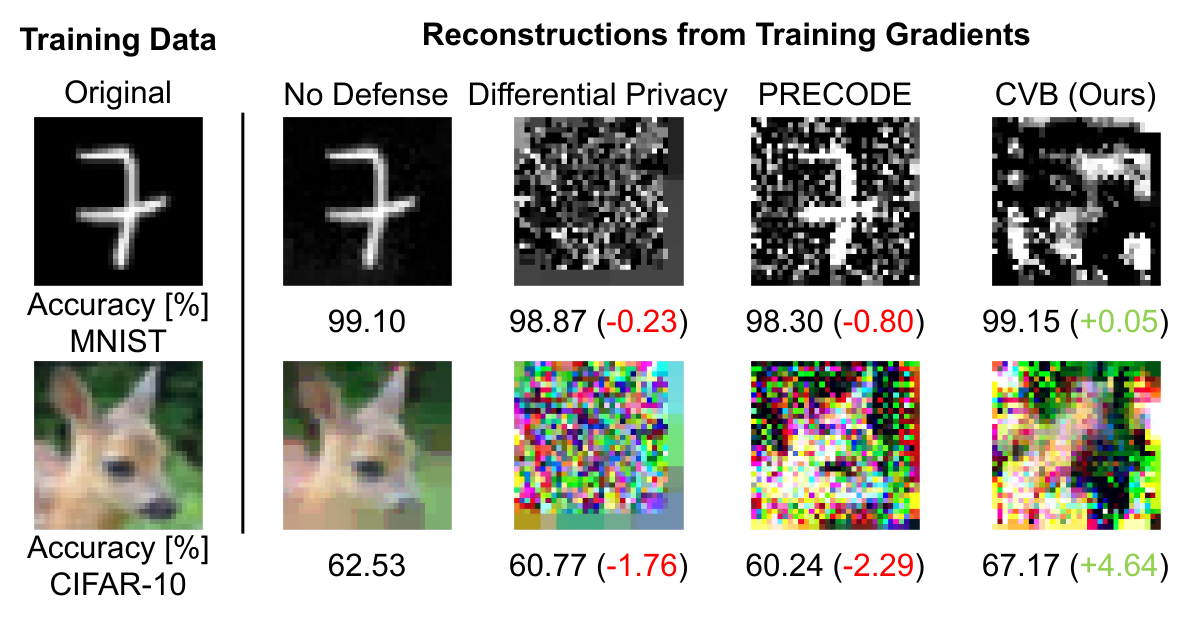}
		\caption{
		\textbf{Content summary of this paper.} 
		Neural networks are trained on the MNIST and CIFAR-10 dataset in a federated scenario.
		As training gradients leak private training data, different defense mechanism can be used for protection.
		While state-of-the-art perturbation techniques such as \textit{Differential Privacy} can prevent reconstruction, they reduce model utility.
		\textit{PRECODE} can preserve privacy if placed early in a model, but at the cost of reduced model utility and increased computational and communication resources.
            We propose a novel \textit{Convolutional Variational Bottleneck} (CVB) to preserve privacy with notably less costs and improved model utility.
		}
		\label{fig:summary}
	\end{center}
\end{figure}

\IEEEpubidadjcol

The de-facto standard defense against such privacy leaks is to perturb the exchanged gradients~\cite{bonawitz2017practical, jayaraman2019evaluating, zhu2019deep, papernot2021tempered, sattler2019robust, jin2020stochastic, wei2021gradient, ponomareva2023dp, wei2023securing}.
However, gradient perturbation results in an inherent trade-off between model utility and privacy~\cite{dwork2014algorithmic, jayaraman2019evaluating, zhu2019deep, wei2020framework, scheliga2022precode, huang2021evaluating, el2022differential, ponomareva2023dp}.
To avoid this trade-off, a model-based defense mechanism called \textit{PRivacy EnhanCing mODulE} (PRECODE), has been proposed as a generic extension for arbitrary model architectures~\cite{scheliga2022precode}.
To obscure the original latent feature space, PRECODE utilizes a variational bottleneck placed within the original model.
The stochasticity introduced by variational modeling is supposed to counter iterative gradient inversion attacks by design.

In this paper, we thoroughly investigate the effect of PRECODE on gradient inversion attacks.
Our analysis shows that the variational modeling introduced by PRECODE primarily affects the stochasticity in the gradients of its subsequent network layers.
Based on this finding, we show that the privacy preserving effect of PRECODE can be disabled by purposefully omitting these stochastic gradients during attack.

Hence, PRECODE must be placed early in a neural network to maintain privacy.
However, the variational bottleneck is based on fully connected layers.
This causes an increase in additional model parameters and thus an increase in computational and communication overhead.
In addition, potentially relevant positional information present in data such as images, is lost when modeling with fully connected layers.

To address these challenges we propose a novel model-based defense mechanism that utilizes fully convolutional variational bottlenecks.
An extensive experimental evaluation shows that our proposed defense mechanism (1) increases privacy against gradient inversion attacks; (2) requires fewer additional parameters and therefore less computational and communication resources than PRECODE; and (3) can improve model utility through a regularization-like effect on the modeled data.
Fig.~\ref{fig:summary} is a visual summary of the content of this paper.
Our contributions can be summarized as follows:
\begin{itemize}
    \item We analyze the model gradients during iterative gradient inversion attacks to reveal how PRECODE counteracts such attacks by design.
    \item We propose an attack that disables the privacy preserving properties of PRECODE by explicitly omitting stochastic gradients from the attack optimization.
    \item We propose a novel privacy module -- the Convolutional Variational Bottleneck (CVB). 
    Our CVB leverages the privacy inducing stochastic effects of variational modeling and simultaneously allows for increased model utility as well as lower computational and communication overhead compared to PRECODE.
    \item We perform a systematic empirical evaluation on three seminal model architectures and six image classification datasets of increasing complexity. 
    We evaluate the robustness of our CVB against three state-of-the-art gradient inversion attacks and compare our method to three state-of-the-art defenses under various configurations.
\end{itemize}

The remainder of this work is structured as follows:
Section~\ref{sec:sota} introduces our threat model and summarizes related work on gradient inversion attacks and defenses.
Thereupon we discuss our analysis of PRECODE regarding the impact on gradients.
Based on this analysis, we formulate our attack to counteract PRECODEs privacy preservation.
Subsequently, we introduce and discuss our proposed CVB defense.
Section~\ref{sec:design} describes the experimental setup used to evaluate the defense methods.
Experimental results and their discussion are presented in Section~\ref{sec:results}.
Section~\ref{sec:conclusion} concludes our work.

\section{Related Work}
\label{sec:sota}

\subsection{Threat Model}
\label{sec:threat_model}

\begin{figure*}[t]
	\begin{center}
		\includegraphics*[width=.99\linewidth]{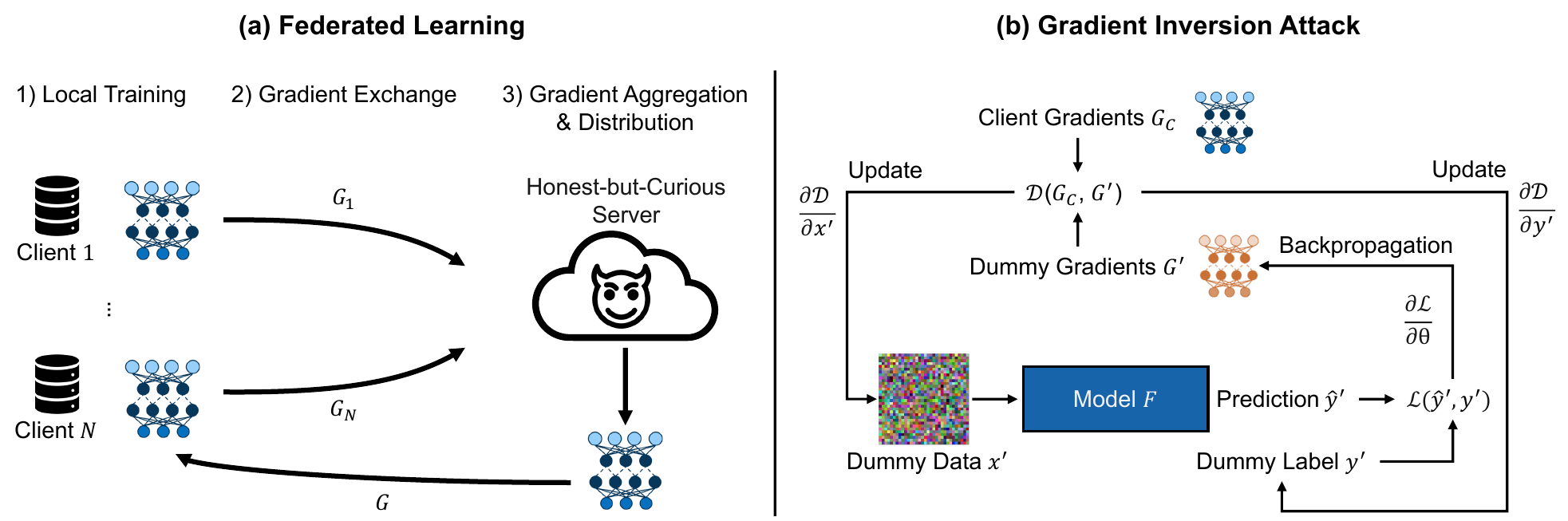}
		\caption{(a) Investigated federated learning setting; (b) Threat model and attack process of gradient inversion attacks.}
		\label{fig:FL}
	\end{center}
\end{figure*}

We investigate the privacy of collaborative training processes such as \textit{Federated Learning} (FL)~\cite{mcmahan2017communication, kairouz2021advances}.
Fig.~\ref{fig:FL}(a) illustrates a typical FL process.
In FL, training data is not shared but remains local with each client.
Clients who participate in the collaborative training process initialize their local model from a global model state.
After a defined number of local training steps, clients return their local model updates, \ie, their model gradients, and a new global model state is computed by aggregating the clients' gradients.
This process is repeated iteratively until convergence or some other termination criterion is satisfied. 
All collaborative training approaches that are based on the exchange of gradient information suffer from similar vulnerabilities, \eg, peer-to-peer or cluster-based collaborative training~\cite{roy2019braintorrent, lee2020tornadoaggregate, duan2020self}.

Consistent with related work, we adopt the \textit{honest-but-curious} server threat model, which is particularly relevant for such collaborative training processes~\cite{zhu2019deep, zhao2020idlg, geiping2020inverting, wei2020framework, wang2020sapag, yin2021see, jin2021cafe, jeon2021gradient, li2022auditing}.
In this scenario the server \textit{honestly} participates in the training process, \ie, it does not actively interfere with it.
In this role, the attacker has knowledge of the training loss function $\mathcal{L}$, the model $F$, the model weights $\theta$, and the exchanged clients training gradients $G_C = \nabla \mathcal{L}_\theta(\hat{y}, y)$.
$\hat{y}=F(x)$ denotes the prediction of the attacked model $F$ after forward propagation of an input $x$, whereas $y$ is the expected output.
The attacker is \textit{curious} in that she leverages this knowledge to compromise client privacy with particularly sophisticated \textit{Gradient Inversion} attacks.

\subsection{Gradient Inversion Attacks}
\label{sec:sota_gi}
Gradient Inversion (GI) attacks aim to reconstruct training data $(x, y)$ from exchanged gradients $\nabla \mathcal{L}_\theta(\hat{y}, y)$~\cite{zhu2019deep}.
Fig.~\ref{fig:FL}(b) illustrates the general attack process.
A randomly initialized dummy sample $x'$ is fed into the model $F$ to obtain a dummy gradient $G' = \nabla \mathcal{L}_\theta(F(x'),y')$, where $y'$ is a randomly initialized dummy label.
To reconstruct the original training data $(x, y)$, the distance $\mathcal{D}$ between the client gradient $G_C = \nabla \mathcal{L}_\theta(F(x), y)$ and the dummy gradient $G'$ is minimized by iteratively updating the dummy data $(x', y')$.
$\Omega$ describes an additive regularization term which is weighted by $\lambda$.
The dummy data is adjusted by gradient based optimization until convergence.
Given these notations, such GI attack can generally be formulated as the following optimization problem:
\begin{equation}
    \label{eq:GI}
    \argmin_{ \left( x',y' \right) } \mathcal{D}(\nabla \mathcal{L}_\theta(F(x), y),  \nabla \mathcal{L}_\theta(F(x'),y')) + \lambda \Omega.
\end{equation}

\textit{Deep Leakage from Gradients} (DLG)~\cite{zhu2019deep} first formulates this iterative GI attack by minimizing the Euclidean distance $\mathcal{D}$ using a L-BFGS optimizer~\cite{liu1989limited} without regularization term.
Zhao \etal improve DLG (iDLG)~\cite{zhao2020idlg} by analytical reconstruction of the ground-truth labels $y$ in advance.
It was found that omitting the need to optimize for $y'$ in Eq.~\ref{eq:GI} accelerates and stabilizes the optimization process.
Although iDLG only guarantees to reconstruct the labels for batches of size $1$, the authors of~\cite{yin2021see} have formulated a method to reliably reconstruct label information for large batches of any size that contain disjoint classes.

The \textit{Client Privacy Leakage} (CPL) attack~\cite{wei2020framework} additionally utilizes a label-based regularizer to enhance the stability of the attack optimization.
Specifically the regularizer minimizes the Euclidean distance between the predicted dummy label $\hat{y}'$ and the analytically reconstructed ground truth label $y$.

The \textit{Inverting Gradient} (IG) attack~\cite{geiping2020inverting} further improves the reconstruction process by disentangling gradient direction and magnitude.
Instead of the Euclidean distance, the cosine distance between client and dummy gradients is used as distance function $\mathcal{D}$ in Eq.~\ref{eq:GI}.
Furthermore, the authors use total variation~\cite{rudin1992nonlinear} of the dummy image $x'$ as an image prior to enhance the fidelity of their reconstructions.
Additionally, the Adam optimizer~\cite{kingma2014adam} was found to generally yield more advanced reconstruction results compared to L-BFGS.
Further work on improving the quality of iterative GI attacks mainly focuses on the choice of: 1) the gradient distance function $\mathcal{D}$, 2) the regularization term $\Omega$, 3) the optimizer and 4) the label reconstruction method~\cite{wei2020framework, wang2020sapag, yin2021see, jin2021cafe, jeon2021gradient}.
An elaborate overview for recent attacks can be found in \cite{li2022survey, zhang2022survey}.

In our work we focus on iterative GI attacks.
We acknowledge the presence of analytical recursion-based GI attacks~\cite{aono2017privacy, fan2020rethinking, zhu2020r}.
However, these are generally limited in terms of gradients that accumulate very few input samples.
Furthermore, they require the attacked neural networks to use bias weights.
Removing the bias weights from all fully connected layers in a neural network makes these attacks infeasible.
More information regarding the impact of bias weights on analytical GI attacks can be found in the \textit{Supplementary Materials}.
In result, iterative GI attacks are the biggest threat for privacy leakage from gradients.

\subsection{Defense Mechanisms}
\label{sec:sota_def}
The authors of~\cite{wei2020framework} present a comprehensive analysis of GI attacks and deduce relevant parameters as well as potential mitigation strategies. 
They found that batch size, image resolution, choice of activation functions and the number of local training epochs can impact gradient leakage.
Supporting findings are reported in \cite{zhu2019deep, zhao2020idlg, geiping2020inverting, pan2022exploring, zhu2020r}.
Furthermore, the impact of training progress on gradient leakage was observed, as models trained for more communication rounds tend to yield smaller gradients compared to earlier stages of the training process~\cite{geiping2020inverting, wang2020sapag, scheliga2022precode}.
However, even if training parameters are carefully selected to prevent GI attacks, there is no guarantee that reconstruction of sensitive data is not possible~\cite{geiping2020inverting}. 
As a matter of fact, \textit{Geiping}~\etal showcased successful attacks on deep neural networks (ResNet-152) trained for multiple communication rounds and batches of $100$ images~\cite{geiping2020inverting}. 
Additionally, parameter selection is often controlled by other factors, like model and/or hardware limitations.
Therefore, to achieve data privacy, defense mechanisms should be actively developed, analyzed and applied.

\subsubsection{Cryptography-based Defense}
Although cryptographic methods such as secure aggregation schemes~\cite{yao1982protocols, bonawitz2017practical} and homomorphic encryption~\cite{gentry2009fully, aono2017privacy, zhang2020batchcrypt} are likely to provide the best privacy preservation, they come at extreme computational costs and are not applicable in every collaborative learning scenario~\cite{el2022differential}.
For example Pasquini \etal~\cite{pasquini2022eluding} demonstrate that for particular FL scenarios, secure multiparty computation schemes can be practically bypassed, allowing for GI attacks as described above. 
Instead of using costly cryptographic methods, the de-facto standard to defend against privacy leakage is perturbation of the exchanged gradients.

\subsubsection{Gradient Perturbation}
Typically three approaches are utilized for gradient perturbation: \textit{gradient quantization}, \textit{gradient compression}, and \textit{noisy gradients}.
\textit{Gradient quantization} primarily aims to reduce communication costs and memory consumption during collaborative training~\cite{konevcny2016federated, sattler2019robust, deng2020model}. 
Fixed ranges of numerical values are compressed to sets of values, reducing the entropy of the quantized gradients.
As a side effect, this also reduces the success of GI attacks.
The authors of~\cite{zhu2019deep} found only low-bit Int-8 quantization to be sufficient for defending against gradient leakage. 
However, this caused a $22.6\%$ drop in model accuracy~\cite{zhu2019deep}, which is not acceptable in practical scenarios. 

\textit{Gradient compression} or \textit{sparsification} through pruning is a method originally used to reduce the communication costs during collaborative training~\cite{lin2017deep, tsuzuku2018variance, deng2020model}. 
For each layer of the model, gradient elements that carry the least information, \ie, have the smallest magnitude, are pruned to zero.
Similar to quantization, this method reduces the reconstruction success for GI attacks at cost of model utility~\cite{zhu2019deep, wei2020framework, huang2021evaluating}.

The concept of using noise to limit information disclosure about individuals was first introduced in the field of differential privacy~\cite{dwork2014algorithmic, abadi2016deep, kairouz2021advances}.
To guarantee a provable degree of privacy to clients participating in collaborative training, state-of-the-art methods use \textit{noisy gradients}~\cite{bonawitz2017practical, li2019privacy, kairouz2021advances, ponomareva2023dp}.
Naive approaches simply add Gaussian or Laplace noise to the gradients prior to their exchange.
More sophisticated approaches use a combination of quantization or gradient clipping and a carefully tuned noise injection~\cite{jin2020stochastic, wei2021gradient}.

A major advantage of differentially private algorithms are the \textit{theoretical guarantees} on how much privacy a model may leak.
Although the addition of noise may suppress GI attacks, it can also negatively affect the training process and final model utility.
Wei~\etal demonstrate that noise must be added at levels that cause model accuracy drops of up to $9.8\%$ to protect the exchanged data from privacy leakage~\cite{wei2020framework}. 
Increasing the amount of training data might mitigate such drop in model utility~\cite{dwork2014algorithmic}.
However, increasing the amount of training data is typically not feasible in practical scenarios.
In addition, Wang~\etal show that the theoretical privacy guarantees of DP do not necessarily guarantee practical privacy~\cite{wang2022differential}.

\subsubsection{Model Extension}
As discussed above, gradient perturbation is the de-facto standard for defending against gradient leakage.
However, gradient perturbation is inherently limited by a well observed trade-off between privacy and model utility~\cite{dwork2014algorithmic, jayaraman2019evaluating, zhu2019deep, wei2020framework, scheliga2022precode, huang2021evaluating}.

In contrast to gradient perturbation, the authors of~\cite{scheliga2022precode} propose to extend neural network architectures with a \textit{PRivacy EnhanCing mODulE} (PRECODE) that aims to protect against GI attacks.
The module can be generically integrated into any existing model architecture without requiring further model modifications.
More importantly, the module does not notably harm the final model utility or training process, \eg, by causing increased convergence times.

\begin{figure}[!t]
	\begin{center}
		\includegraphics*[width=.99\linewidth]{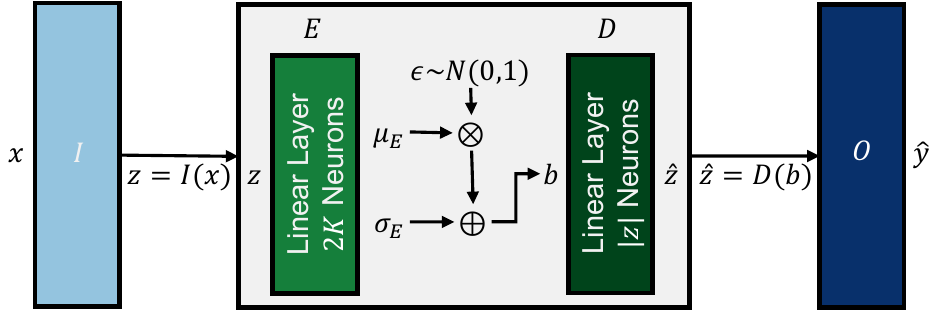}
		\caption{Realization of the PRECODE extension as variational bottleneck.} 
		\label{fig:vb}
	\end{center}
\end{figure}

PRECODE is implemented as a variational bottleneck (VB), which originates from the field of generative modeling.
VBs aim to learn a joint distribution between the input data and its latent representation~\cite{kingma2013auto}.
As visualized in Fig.~\ref{fig:vb}, they consist of a probabilistic encoder $E=q(b|z)$ and decoder $D=p(z|b)$. 
The VB approximates the distribution of the latent space and obtains new representations $\hat{z}$ from an approximated prior distribution $p(b)$ by stochastic sampling from a multivariate Gaussian.

The encoder $E$ is realized as a fully connected layer with $2K$ neurons, where $K$ defines the size of the bottleneck representation $b$.
$z=I(x)$ represents the latent representations computed by forward propagating an input sample $x$ through all layers $I$ of the base neural network prior to the output layer.
The bottleneck $E$ encodes this representation $z$ into a latent distribution which fits a multivariate Gaussian: 
\begin{equation}
    \label{eq:bottleneck}
    E(z) = q(b|z) = \mathcal{N}(\mu_E, \sigma_E)
\end{equation}
where $\lbrack \mu_E, \sigma_E \rbrack = \lbrack \mu_1, \mu_2,\dots, \mu_K, \sigma_1, \sigma_2,\dots, \sigma_K \rbrack$ is the output of encoder neurons.

The bottleneck features $b \sim q(b|z)$ are then fed into the stochastic decoder $D$, which is also realized as a fully connected layer. 
The number of neurons in $D$ matches the size of $z$.
$D$ computes a new latent representation $\hat{z}=p(z|b)p(b)$ which is eventually used to calculate the model prediction $\hat{y} = O(\hat{z})$, where $O$ corresponds to the remaining layers of the base model behind the VB.

The loss function $\mathcal{L}_F$ of the base model $F$ is extended by the Kullback-Leibler divergence ($\mathcal{D}_{KL}$) between $\mathcal{N}(\mu_E, \sigma_E)$ and a standard normal distribution, so that the VB learns a complete and continuous latent feature space distribution:
\begin{equation}
    \label{eq:vb_loss}
    \mathcal{L}(\hat{y}, y) = \mathcal{L}_F(\hat{y},y) + \beta \cdot \mathcal{D}_{KL}(\mathcal{N}(\mu_E, \sigma_E), \mathcal{N}(0, 1)).
\end{equation}
$\beta$ controls the weight of the VB loss on the overall loss function.
To backpropagate a gradient through the bottleneck layer, the reparameterization technique described in~\cite{kingma2013auto} is used.
By default, PRECODE is inserted between the last feature extraction layer and the output layer of a given base neural network.

\section{Decoding PRECODE}
\label{sec:analysis}
The VB realizes a \textit{stochastic function} which transforms input representations $z$ to stochastic latent representations $\hat{z}$.
As $\hat{y}$ and $\mathcal{L}(\hat{y}, y)$ are computed based on this stochastic representation, the gradient $\nabla \mathcal{L}(\hat{y}, y)$ does not contain direct information on the input sample $x$.
Furthermore, iterative optimization-based attacks are countered by design, since $\hat{z}'=D(E(I(x')))$ is influenced by random sampling in each optimization step during an attack.
Even small changes in $x'$ cause an increased entropy of $\hat{z}'$.
This makes it difficult for the optimizer to find dummy images $x'$ that minimize the reconstruction loss.

\subsection{Gradient Analysis}
\begin{figure}[!t]
	\begin{center}
		\includegraphics*[width=1\linewidth]{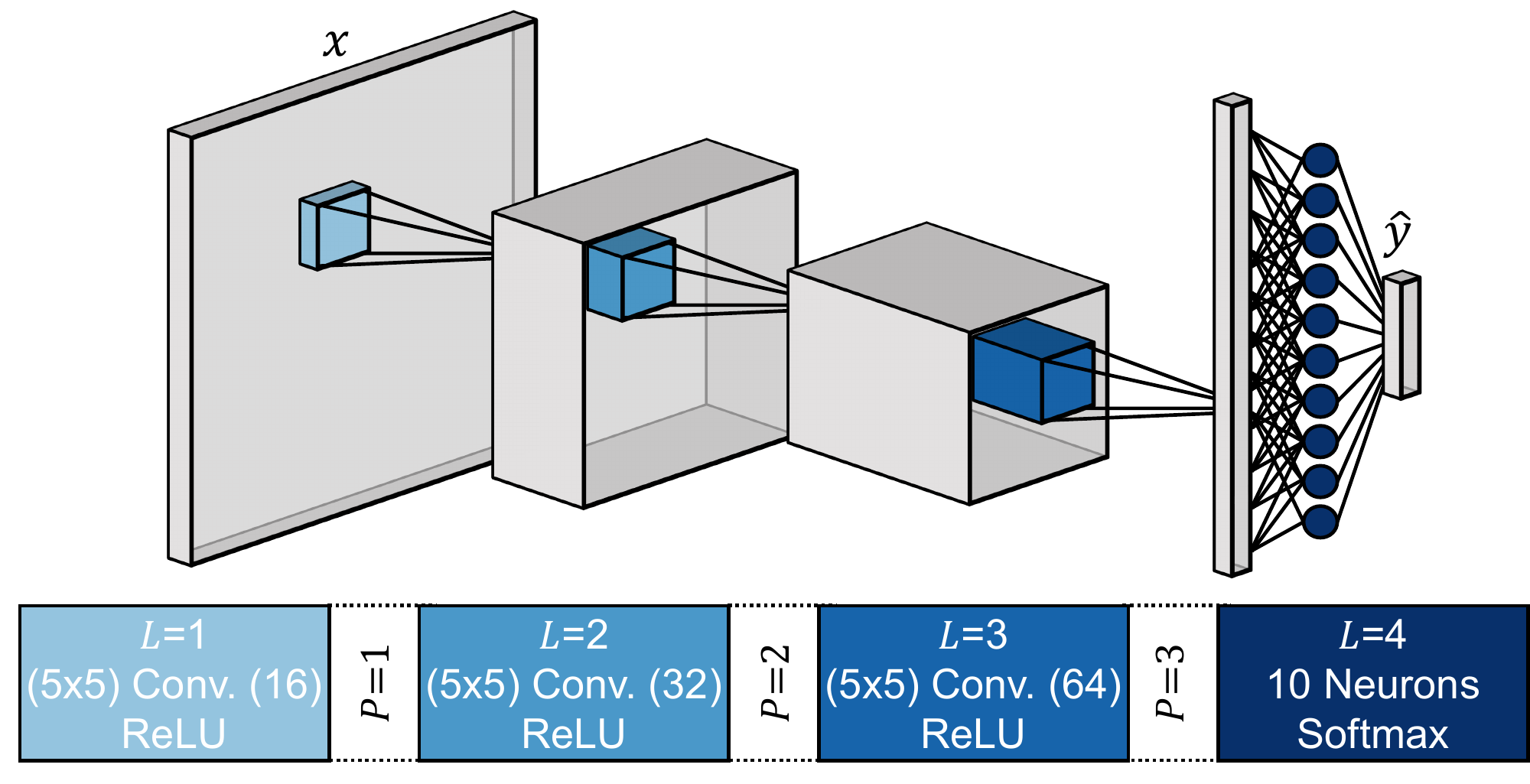}
		\caption{Architecture of the baseline CNN used in the experiments of this paper. $P$ indicates the position at which privacy modules can be placed.}
		\label{fig:cnn}
	\end{center}
\end{figure}
To analyze the impact of PRECODE on the models' gradients during an iterative GI attack, we tracked the evolution of dummy gradients for each iteration of an attack, \ie, for each update of the dummy image $x'$.
The same was done for the model without PRECODE.
We realized a very basic attack scenario for our gradient analysis, that uses a small convolutional neural network (CNN), a single random image from the CIFAR-10 dataset~\cite{krizhevsky2009learning} and IG~\cite{geiping2020inverting} as GI attack.
The model architecture is illustrated in Fig.~\ref{fig:cnn}.
The CNN has three convolutional layers with $5\times 5$ kernels, $\lbrack 16, 32, 64 \rbrack$ channels, a stride of $2$ and ReLU activation.
The last layer of the CNN is a fully connected classification layer with ten neurons and softmax activation.
As suggested in~\cite{scheliga2022precode} we place PRECODE at position $P=3$ between the last convolutional layer and the classifier.
We repeated this analysis for different model inputs and observed similar behaviour.
More details on the experimental setup can be found in Section \ref{sec:design} and the \textit{Supplementary Material}.

\begin{figure*}[!t]
\centering
\subfloat[]{\includegraphics[height=.11\textheight]{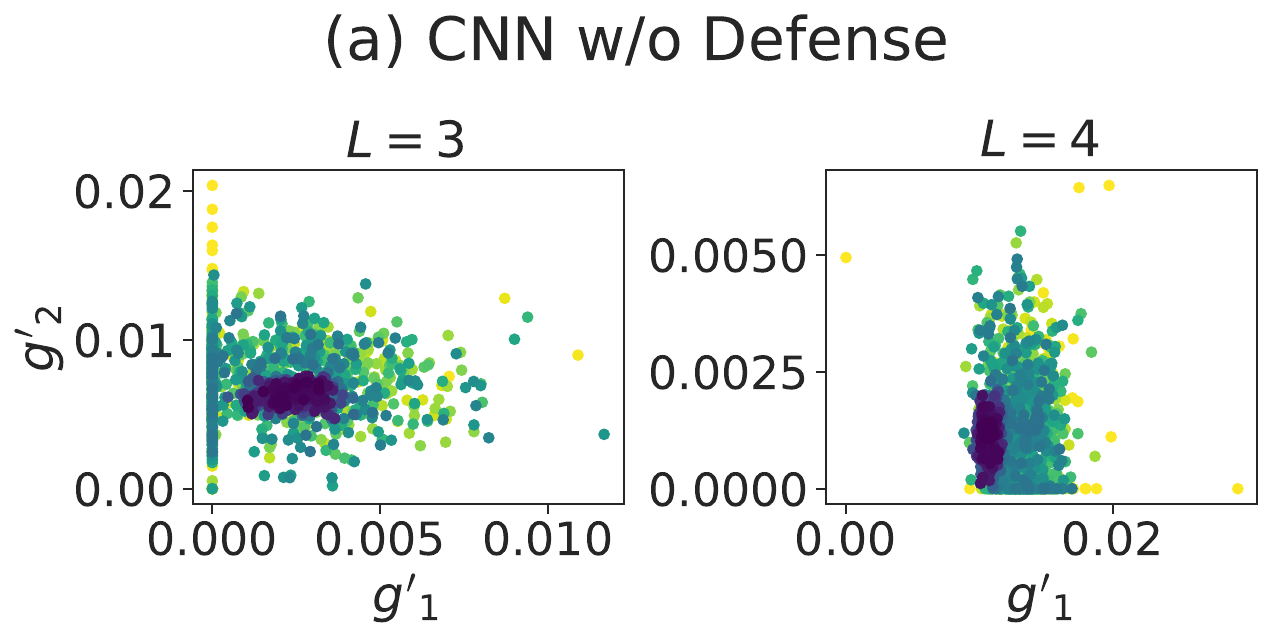}%
}
\hfil
\vrule
\hfil
\subfloat[]{\includegraphics[height=.11\textheight]{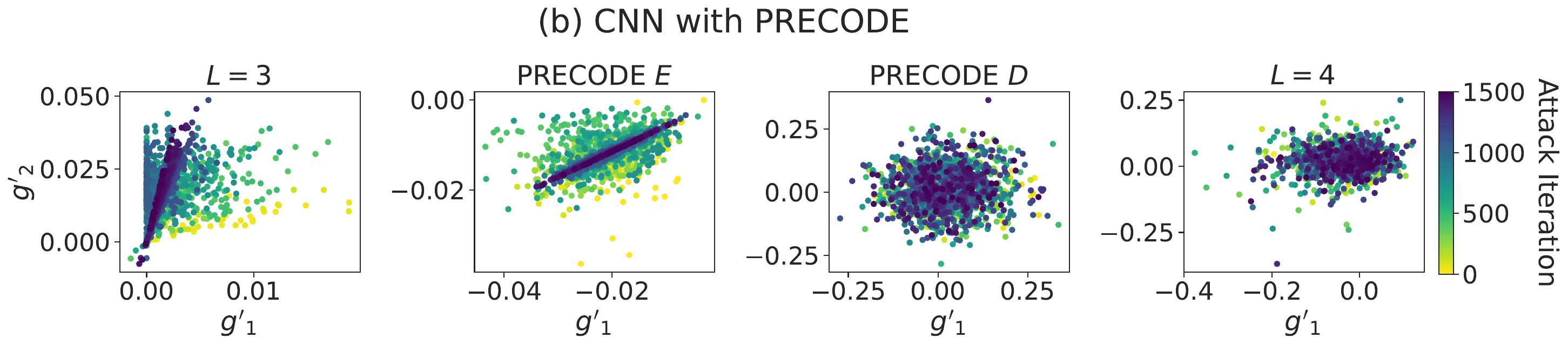}%
}
\caption{\textbf{Behaviour of dummy gradients during a GI attack} for (a) a CNN and (b) a CNN that is protected by PRECODE. Two random gradient values $g'_1$ and $g'_2$ of dummy gradients $G'$ of the indicated layers are tracked over the course of the GI attack. Color represents the attack iteration.}
\label{fig:attack_gradients}
\end{figure*}

\begin{figure*}[htb]
\centering
\includegraphics[width=\textwidth]{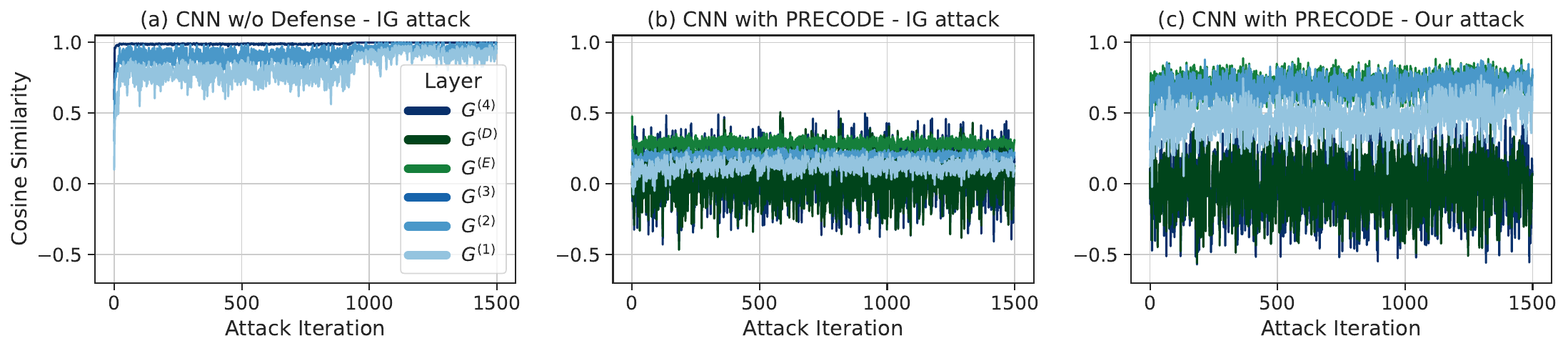}
 \caption{\textbf{Behaviour of dummy gradients during a GI attack.} 
 Cosine similarities between dummy gradients and the client gradient are tracked over the course of the GI attack for (a) a CNN without protection and (b-c) the CNN protected by PRECODE. 
 (a) and (b) are attacked with IG~\cite{geiping2020inverting}; (c) is attacked with our proposed "Ignore" attack described in Section~\ref{sec:anti_precode}. 
 Color represents the layer of the model. CNN layers use blue shades and the PRECODE VB layers use green ones.}
\label{fig:precode_cossim}
\end{figure*}

Fig.~\ref{fig:attack_gradients} shows the change of two randomly chosen dummy gradient values $g'_1$ and $g'_2$ of different layers of the model.
Specifically (a) shows the gradient values for the last convolutional layer ($L=3$) and the final classification layer ($L=4$) of the CNN.
Additionally (b) shows the behavior of the gradient values of the PRECODE module ($E$ and $D$) for the same CNN protected by PRECODE.
The iteration progress is represented by color change.

For the unprotected baseline model in (a), the dummy gradient values converge towards the corresponding client gradient values.
This in turn indicates that the dummy image $x'$ converges to an image that leads to a similar gradient as the original client training data $x$.
For the PRECODE protected model the corresponding effects on the dummy gradients can be observed in Fig.~\ref{fig:attack_gradients} (b).
The gradient values of layers after stochastic sampling, \ie, the PRECODE decoder ($D$) and classification layer ($L=4$), fluctuate widely.
The gradient of the previous layers are also affected by stochastic sampling.
However, they tend to be more stable.

Fig.~\ref{fig:precode_cossim}(a-b) further illustrate this behaviour in an aggregated manner.
In (a), the cosine similarities between the dummy and client gradients for all layers of the baseline CNN increase rapidly in the first few iterations of the attack, as the dummy images and their gradients converge towards the original ones.
Also, the cosine similarities for the gradients of the later layers are generally larger.

For the PRECODE protected model in Fig.~\ref{fig:precode_cossim}(b), the cosine similarities for the decoder and classifier fluctuate widely at a low similarity level.
Since the gradient direction for these layers changes with each attack iteration (\cf Fig.~\ref{fig:attack_gradients}(b)), the optimization process cannot converge.
Although more stable, the cosine similarities are even lower for earlier layers.
For the first layer they even constantly remain near zero.

The random sampling process in the PRECODE bottleneck causes the latent representation of the dummy image $\hat{z}'$ to take new values in each attack iteration.
Even if the same dummy image $x'$ is forwarded through the model, stochastic sampling in PRECODE would result in different dummy gradients.
Hence, if the dummy gradient points into a different direction at every attack iteration, the optimizer of the GI attack cannot find a consistent direction to minimize the distance between the dummy and the client gradient.
Similar to our observations, recent work~\cite{scheliga2023dropout} also found that stochastic gradients lead to poor reconstructions, because the attack optimization process does not converge due to unstable dummy gradients.

\subsection{Attacking PRECODE}
\label{sec:anti_precode}

From the above analysis we learned that (1) stochastic gradients lead to poor reconstructions and (2) the largest dummy gradient fluctuations during the attack can be observed in layers after stochastic sampling of the PRECODE decoder.
Balunovic~\etal showed that the impact of defense mechanisms can be reduced or even removed by purposefully ignoring specific parts of a defense mechanism~\cite{balunovic2021bayesian}.
Therefore, and based on our findings, we propose 
to ignore stochastic gradients during attack optimization.
Given the gradients $\nabla \mathcal{L}_\theta(F(x), y) = G = \lbrack G^{(1)}, \dots, G^{(E)}, G^{(D)}, \dots, G^{(L)} \rbrack$ we omit $\lbrack G^{(D)}, \dots, G^{(L)}  \rbrack$, \ie the gradients of PRECODE's decoder and the subsequent layers, during the attack. 
Superscripts $L$ in $G^{(L)}$ denote the layers position (\cf Fig.~\ref{fig:cnn}) and $E$ and $D$ denote PRECODE's encoder and decoder layer.
Correspondingly $G' = \nabla \mathcal{L}_\theta(F(x'), y')$ refers to the dummy gradient.
Hence the reconstruction loss for our GI attack is only calculated through the distance $\mathcal{D}$ between the client and dummy gradients of those layers $\lbrack G^{(1)}, \dots, G^{(E)} \rbrack$, that precede the PRECODE decoder $D$:
\begin{equation}
    \label{eq:precode_att}
    \mathcal{D}(\lbrack G^{(1)}, \dots, G^{(E)} \rbrack, \lbrack G'^{(1)}, \dots, G'^{(E)} \rbrack)
\end{equation}
The adjustments on the dummy image would stabilize as they are based only on the more stable dummy gradients.
Hence, the dummy data could converge towards the actual input data.

We display the empirical evaluation of our "Ignore" attack in Fig.~\ref{fig:precode_cossim}(c).
Although the dummy gradients still fluctuate widely, the cosine similarities between dummy and client gradients increase for the layers preceding the PRECODE decoder.
This indicates that the dummy gradients converge towards the client gradients and that original training data can be reconstructed.
The cosine similarities of the PRECODE decoder $D$ and classification layer ($L=4$) remain as low as for the IG attack.
However, they do not impact the optimization process anymore.
Backpropagation through the VB increases the variance of the dummy gradients.
In result, whereas the attack optimization of the baseline model quickly converged to cosine similarities close to $1$ (\cf Fig.~\ref{fig:precode_cossim}(a)), the dummy gradients with PRECODE protection only achieve similarities between $0.7$ and $0.8$ (\cf Fig.~\ref{fig:precode_cossim}(c)).

\begin{table}
\centering
\caption{
Attack Success Rate (ASR) for attacked gradients of a CNN for $128$ images of the CIFAR-10 dataset.
We attack different combinations of gradients as indicated by \cmark and \xmark. 
"-" indicate the baseline CNN without defense, as it does not use the PRECODE VB layers $E$ and $D$.
Colors in the table head represent the layer of the model, whereas CNN layers use blue shades and PRECODE VB layers use green ones.
}
\label{tab:attack_comparison}
\resizebox{\linewidth}{!}{%
\begin{tabular}{c|cccccc|r}
\toprule

       & \multicolumn{6}{c|}{Attacked Gradient}               &  \\
Attack & \color[HTML]{F1F1F1} {\cellcolor[HTML]{94c4df}} $G^{(1)}$ & \color[HTML]{F1F1F1} {\cellcolor[HTML]{4a98c9}} $G^{(2)}$ & \color[HTML]{F1F1F1} {\cellcolor[HTML]{1764ab}} $G^{(3)}$ & \color[HTML]{F1F1F1} {\cellcolor[HTML]{157f3b}} $G^{(E)}$ & \color[HTML]{F1F1F1} {\cellcolor[HTML]{00441b}} $G^{(D)}$ & \color[HTML]{F1F1F1} {\cellcolor[HTML]{08306b}} $G^{(4)}$ & ASR $[\%]\downarrow$\\
\midrule
\multirow{2}{*}{IG} & \cmark & \cmark & \cmark & - & - & \cmark &    \textbf{96.88} \\
 & \cmark & \cmark & \cmark & \cmark & \cmark & \cmark &   0 \\
\midrule

\multirow{3}{*}{Ignore} & \cmark & \cmark & \cmark & \cmark & \xmark & \cmark &   0 \\
 & \cmark & \cmark & \cmark & \cmark & \cmark & \xmark &   0 \\
 & \cmark & \cmark & \cmark & \cmark & \xmark & \xmark &    \textbf{85.94} \\

\midrule
\midrule
Attack & \color[HTML]{F1F1F1} {\cellcolor[HTML]{94c4df}} $G^{(1)}$ & \color[HTML]{F1F1F1} {\cellcolor[HTML]{4a98c9}} $G^{(2)}$ & \color[HTML]{F1F1F1} {\cellcolor[HTML]{157f3b}} $G^{(E)}$ & \color[HTML]{F1F1F1} {\cellcolor[HTML]{00441b}} $G^{(D)}$ & \color[HTML]{F1F1F1} {\cellcolor[HTML]{1764ab}} $G^{(3)}$ & \color[HTML]{F1F1F1} {\cellcolor[HTML]{08306b}} $G^{(4)}$ & ASR $[\%]\downarrow$\\
\midrule
IG & \cmark & \cmark & \cmark & \cmark & \cmark & \cmark &   0 \\
\midrule
\multirow{5}{*}{Ignore} & \cmark & \cmark & \cmark & \cmark & \cmark & \xmark &   0 \\
 & \cmark & \cmark & \cmark & \cmark & \xmark & \cmark &   0 \\
 & \cmark & \cmark & \cmark & \xmark & \cmark & \cmark &   0 \\
 & \cmark & \cmark & \cmark & \xmark & \xmark & \cmark &   0 \\
 & \cmark & \cmark & \cmark & \xmark & \xmark & \xmark &     \textbf{7.81} \\ 
\bottomrule
\end{tabular}
}
\end{table}

Excluding stochastic gradients from attack optimization weakens PRECODE's intended privacy inducing effects.
Note that it is not sufficient to only ignore the gradients of the PRECODE decoder \textit{or} the classifier.
Instead, the gradients of the decoder \textit{and all} subsequent layers have to be ignored.
To validate this assumption, we specifically ignore only the gradients of the PRECODE decoder \textit{or} the classifier. 
Also, we move PRECODE to an earlier position in the model, \ie, between the second and third convolutional layer ($P=2$). 
The results are displayed in Tab.~\ref{tab:attack_comparison}.
Only if the gradients of the PRECODE decoder \textit{and all} subsequent layers are ignored, the attacker is able to reconstruct data.
Note that the ASR reduces to $85.94\%$.
In all other cases the ASR drops to $0\%$.
If we move PRECODE to an earlier position, more gradients are affected by the stochastic sampling.
Hence, less gradient information is usable for our "Ignore" attack, which causes a further reduction of ASR to $7.81\%$.

\subsection{Early PRECODE Placement}
\label{sec:early_precode}
From the analysis above, we conclude that PRECODE generally increases the difficulty of iterative GI attacks, because it decreases the amount of gradient information that can be used during attack optimization.
Primarily the layers after the PRECODE VB are influenced by stochastic sampling.
Therefore, the module should be placed early in the neural network.
However, early placement of PRECODE comes with two major drawbacks.
First, PRECODE uses fully connected layers.
Hence, early placement of PRECODE leads to a loss of positional information potentially present in the data.
In turn, this can result in decreased model utility.
Second, PRECODE is fully connected to its' input.
Hence, the number of additional parameters required by PRECODE grows with the size of the latent features that are input to the module.
This in turn increases communication costs as well as computational effort during federated learning.

\begin{table}
\centering
\caption{
Privacy and model utility metrics as well as number of parameters for a CNN trained with and without PRECODE. 
Arrows indicate direction of improvement. 
Bold and italic formatting highlight best and worst results respectively.
}
\label{tab:param_comparison}
\resizebox{\linewidth}{!}{%

\begin{tabular}{cc|rrrc}

\toprule
& & SSIM $\downarrow$ & ASR $[\%]\downarrow$ & Accuracy $[\%]\uparrow$ & Parameters ($+[\%]$) $\downarrow$ \\
\midrule
& Baseline & $0.87 \pm 0.11$ & 96.88 & $62.53 \pm 0.16$ & 65962 (0.00) \\
\midrule
\multirow{3}{*}{\rotatebox{90}{Position}} & 3 & \textit{0.63} $\pm$ \textit{0.12} & \textit{85.94} & $62.74 \pm 1.07$ & \textbf{72106} \textbf{(9.31)} \\
& 2 & \textbf{0.23} $\pm$ \textbf{0.14} & $7.81$ & \textbf{62.80} $\pm$ \textbf{0.77} & 104362 (58.22) \\
& 1 & $0.30 \pm 0.09$ &  \textbf{0.78} & \textit{60.24} $\pm$ \textit{0.90} & \textit{141226} \textit{(114.1)} \\
\bottomrule
\end{tabular}

}
\end{table}

Tab.~\ref{tab:param_comparison} illustrates these drawbacks for a CNN trained with FedAvg~\cite{mcmahan2017communication} on the CIFAR-10 dataset with $10$ clients.
We report mean and standard deviation of test accuracies over three different seeds.
More details on the experimental setup can be found in Section~\ref{sec:design} and the \textit{Supplementary Material}.

\begin{figure}
    \centering
    \includegraphics[width=.9\linewidth]{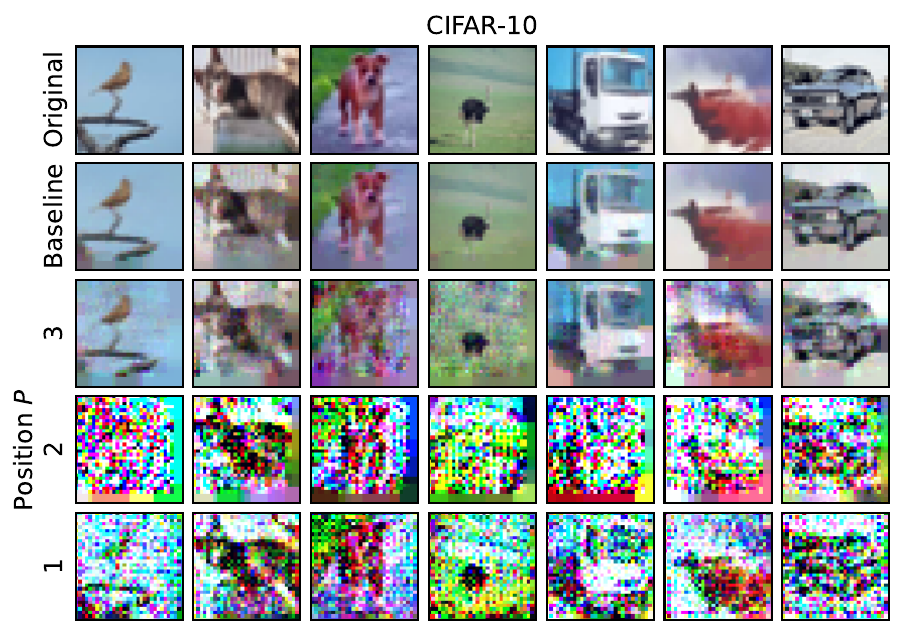}
     \caption{\textbf{Example reconstructions} for a CNN without defense and with PRECODE placed at different positions $P$ for the CIFAR-10 dataset.}
     \label{fig:positions_rec}
\end{figure}

If PRECODE is placed between the last feature extracting layer and the classifier ($P=3)$, as intended in the original publication~\cite{scheliga2022precode}, the model utility in terms of accuracy increases by $1.28\%$ compared to the baseline model without PRECODE.
Similar regularizing effects when extending architectures by variational modeling have been reported in~\cite{alemi2016deep, hofmann2021synaptic}.
The number of trainable parameters only increases by $9.31\%$.
However, as evaluated in Section~\ref{sec:anti_precode}, privacy preservation is insufficient with this placement (\cf Tab.~\ref{tab:attack_comparison}).
If the aim is to preserve privacy, \ie, ASR close to $0\%$, PRECODE must be placed after the first convolutional layer $(P=1)$.
In this case accuracy decreases by $2.12\%$ and the number of model parameters is more than doubled.
Example reconstructions are displayed in Fig.~\ref{fig:positions_rec}.

\section{Convolutional Variational Bottleneck}
\label{sec:convvb}
In the previous section we observed that PRECODE can protect against gradient leakage because stochastic sampling renders gradients of layers after the variational bottleneck unusable for iterative GI attacks.
Hence, to effectively protect against GI, stochastic sampling must take place early in the model.
However, early placement of PRECODE decreases model utility, and increases communication overhead and computational costs.

To alleviate these problems, we propose a novel model extension: the Convolutional Variational Bottleneck (CVB).
Instead of using fully connected layers, our CVB uses convolutional kernels to model the posterior distribution of the latent feature space.
Fig.~\ref{fig:cvb} illustrates our proposed CVB.
We utilize two convolutional kernels $E_\mu$ and $E_\sigma$ of size $(k_E, k_E, K_E)$, which map the the latent features $z=I(x)$ to a mean map $\mu_E$ and a standard deviation map $\sigma_E$ respectively.
Hence, each value in these three-dimensional maps is a statistical representation of one receptive field in the input featuremap $z$.
We utilize the reparameterization trick~\cite{kingma2013auto} to generate samples from the resulting posterior distribution, \ie, $b=\mu_E+\sigma_E \cdot \epsilon$, where $\epsilon \sim \mathcal{N}(0, 1)$.
These bottleneck features $b$ are then fed to the stochastic decoder $D$.
We realize $D$ as a $(1\times 1)$ convolution to calculate the stochastic feature representation $\hat{z}=D(b)$.
We purposefully maintain the shape between $z$ and $\hat{z}$ to allow for seamless integration of our CVB at any position in the model.
In detail, zero-padding is used for the bottleneck convolutions $E_\mu$ and $E_\sigma$ to maintain the spatial dimensions of the input feature maps $z$.
In addition, the number of kernels $c$ in $D$ is equal to the number of channels of $z$.
Ultimately, the stochastic features $\hat{z}$ are fed into the remainder of the network $O$ to calculate model predictions $\hat{y} = O(\hat{z})$.
Consistent with Eq.~\ref{eq:vb_loss} the loss function is extended by the Kullback-Leibler divergence between $\mathcal{N}(\mu_E, \sigma_E)$ and $\mathcal{N}(0, 1)$, so that the CVB learns a complete and continuous latent feature space distribution.
The use of convolutional layers for the variational bottleneck offers three main benefits:

\begin{figure}[!t]
	\begin{center}
		\includegraphics*[width=.99\linewidth]{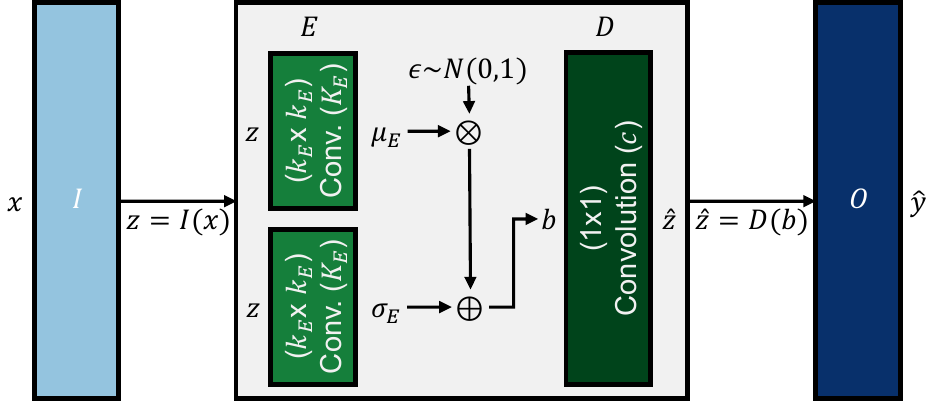}
		\caption{Visualization of our proposed Convolutional Variational Bottleneck (CVB).}
		\label{fig:cvb}
	\end{center}
\end{figure}

\textbf{Parameters:}
First, local connectivity and parameter sharing of convolutions requires fewer model parameters in the VB.
This results in less communication and computational overhead.
The number of parameters for the CVB amounts to $2(k_E^2\cdot c \cdot K_E) + (K_E \cdot c)$.
In the case of fully connected layers, the number of parameters is $2(hwc \cdot K) + (K \cdot hwc)$.
Using our CVB, the number of parameters is independent of the height $h$ and width $w$ of feature maps $z$ that are input to the bottleneck encoder $E$.
Instead, the bottleneck size can be further adjusted via the kernel size $k_E$, which is typically rather small, \ie, between $1$ and $5$.
This provides greater control over the number of parameters added to the model to achieve an optimal trade-off between privacy and utility.
Consider the CNN from the previous section (\cf Fig.~\ref{fig:cnn}), with $k_E=(3,3,3)$, $K = K_E=(8,16,32)$ with PRECODE at position $P=1$.
For input data of shape $(32\times 32 \times 3)$, PRECODE results in $114.10\%$ additional model parameters.
In comparison, our CVB only requires $3.68\%$ additional parameters for the same position.

\textbf{Privacy:}
Secondly, as the bottleneck can be relocated to earlier positions in the model, the stochastic effects that protect from iterative GI attacks can be utilized more effectively.
Earlier placement results in more stochastic gradients and less usable gradient information for the attacker.
Furthermore, local connectivity and parameter sharing of sliding convolutions can offer better protection against gradient leakage compared to fully connected layers.

\textbf{Utility:}
Third, we hypothesize that our CVB can further improve model utility as other related work also observed regularizing effects when extending different architectures with variational modeling~\cite{alemi2016deep, hofmann2021synaptic}.
Furthermore, the convolutions in our CVB preserve spatial information inherent in the feature maps of CNN-based model architectures and may therefore maintain higher model utility.
Additionally, we show how CVB can also be applied to non-CNN-based architectures such as a Vision Transformer (ViT)~\cite{dosovitskiy2020image} to increase privacy and model utility.
We provide a PyTorch implementation of CVB on GitHub\footnote{Published upon acceptance.}.

\section{Experimental Design}
\label{sec:design}
This section describes the general experimental settings used to conduct systematic empirical studies that evaluate model utility and privacy leakage for different hyperparameter choices, gradient inversion attacks, defense mechanisms, datasets and model architectures.

\subsection{Model Utility}
We train models for image classification on the MNIST~\cite{lecun1998gradient} and CIFAR-10~\cite{krizhevsky2009learning} datasets.
The datasets are first separated into training and test splits according to the corresponding benchmark protocols.
We embedded all experiments in a FL scenario with $10$ clients.
The training data splits are independent and identically distributed to those $10$ clients.
Each client creates a validation split that corresponds to $10\%$ of the training data.
This leaves every client with $5'400$/$4'500$ training samples, $600$/$400$ validation samples, and $1'000$/$1'000$ test samples for MNIST/CIFAR-10 respectively.
If we consider other datasets, the same distribution protocol is applied.

The clients collaboratively train a randomly initialized model for $300$ communication rounds using the Federated Averaging algorithm~\cite{mcmahan2017communication}.
In each communication round, (1) a central aggregator transmits the current global model state, (2) each client trains a local model based on the transmitted global model state and the private local training data for $1$ epoch, and (3) the central aggregator aggregates the model gradients of all clients and computes an updated global model state.
Local training minimizes the cross-entropy loss using Adam optimizer~\cite{kingma2014adam} with a learning rate or $0.001$, momentum parameters of $(\beta_1, \beta_2)=(0.9, 0.999)$, and a batch size of $64$.
If PRECODE or CVB is used, the loss function is adjusted according to Eq.~\ref{eq:vb_loss}.
To save computational resources, we stop the training early, if the mean validation loss over all clients has not improved for $40$ consecutive communication rounds.

To measure model utility, accuracy of the global model states is evaluated on the test data after every communication round.
We repeated each experiment and report the mean and standard deviation across three runs with different random seeds.

\subsection{Privacy}
Consistent with related work~\cite{zhu2019deep, zhao2020idlg, geiping2020inverting, wei2020framework, wang2020sapag, yin2021see, jin2021cafe, jeon2021gradient, li2022auditing}, our threat model assumes an \textit{honest-but-curious} attacker (\cf Section~\ref{sec:sota_gi}).
We aim to facilitate reconstruction for the attacker in order to identify an upper bound for privacy leakage via GI attacks.
To evaluate privacy leakage, a \textit{victim dataset} composed of $128$ images is randomly sampled from the training data of one client for each dataset.
A victim gradient is then computed by performing a single training step using that victim data.
To facilitate reconstruction for the attacker, each victim image is attacked independently.
We attack gradients of the models without defense using IG~\cite{geiping2020inverting}.
When PRECODE or our CVB defense is used, we apply our proposed "Ignore" attack (\cf Section~\ref{sec:anti_precode}).
We configured both attacks as follows: dummy images are initialized from a Gaussian distribution, cosine distance is used as loss function, and total variation as regularization term with weight $\lambda_{\text{TV}}=0.01$.
For each attack, Adam optimizer with initial learning rate $1$ is used.
The learning rate is reduced by a factor of $0.1$ if the reconstruction loss plateaus for $400$ attack iterations.
To save computational resources, attacks are stopped if either the reconstruction loss falls below a value of $10^{-5}$, or there is no decrease in reconstruction loss for $4'000$ iterations, or after a maximum of $20'000$ iterations.
To further facilitate reconstruction, we assume that the label information for each attacked sample is generally known.
Please note that label information can be analytically reconstructed from gradients of cross-entropy loss functions \wrt weights of fully connected layers with softmax activation~\cite{geiping2020inverting, zhao2020idlg, wei2020framework}.
This analytical reconstruction of label information might be hindered by the adjusted loss function when using PRECODE or CVB.
However, to avoid systematic advantage of PRECODE and CVB, we assume label information to be generally known irrespective of the defense mechanism.

To measure the reconstruction quality, we compute \textit{Structural Similarity} (SSIM)~\cite{wang2004image} between original and reconstructed images and report the mean and standard deviation of the SSIM across the respective $128$ samples of the victim dataset.
We also compute the Attack Success Ratio (ASR)~\cite{wagner2018technical}, whereas an attack is considered successful if a SSIM of at least $0.5$ was reached.
Privacy leakage correlates with high reconstruction quality and is indicated by high SSIM and ASR values.
The \textit{Supplementary Material} includes further typical metrics such as MSE, PSNR and LPIPS.

\section{Experimental Results}
\label{sec:results}
\subsection{Hyperparameter Study}

\begin{figure*}[!t]
\centering
\subfloat[]{\includegraphics[height=.15\textheight]{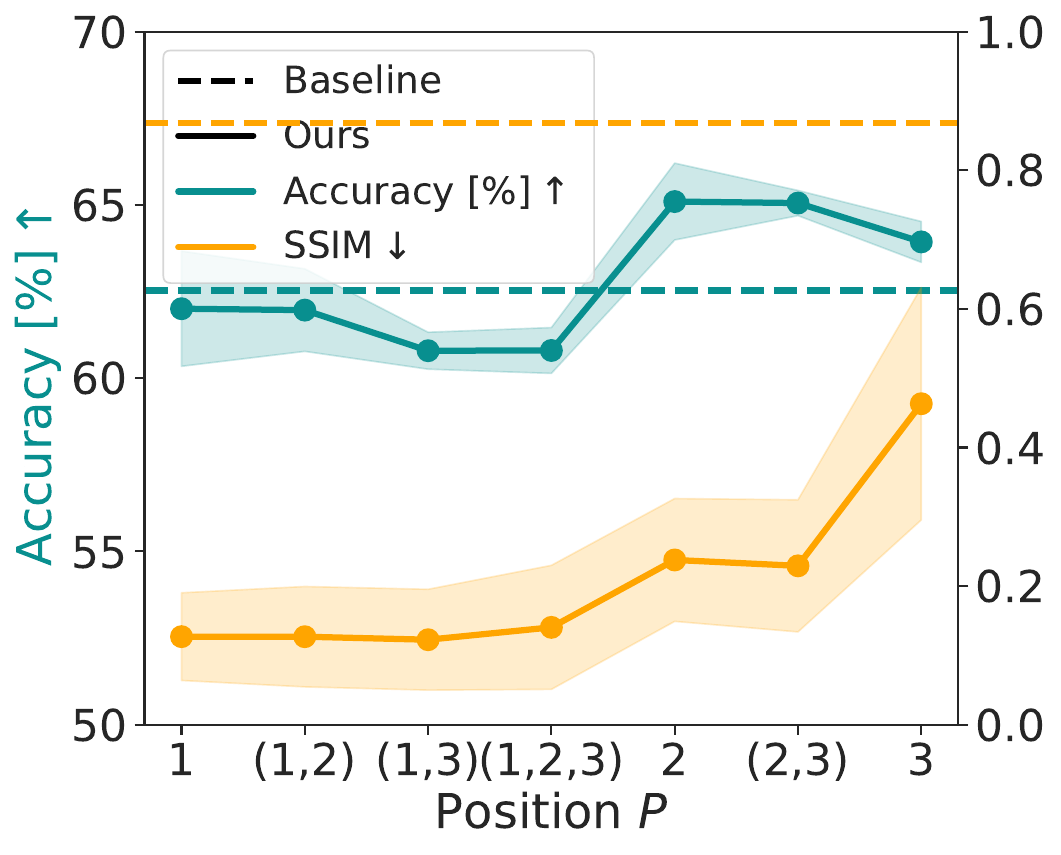}}
\hfil
\subfloat[]{\includegraphics[height=.15\textheight]{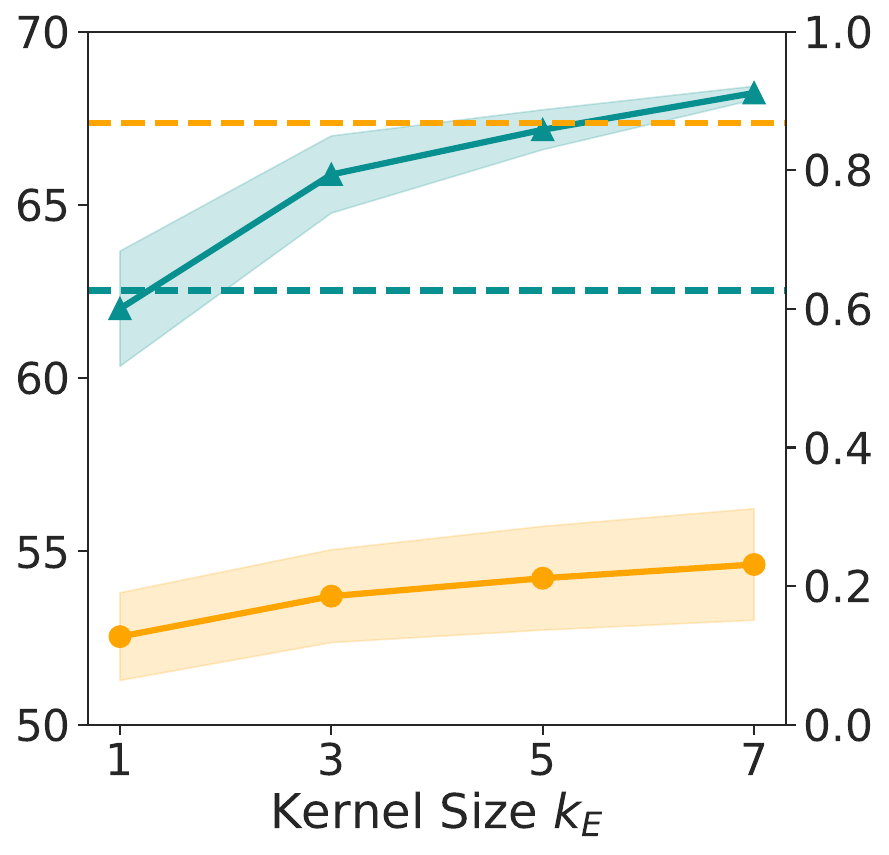}}
\hfil
\subfloat[]{\includegraphics[height=.15\textheight]{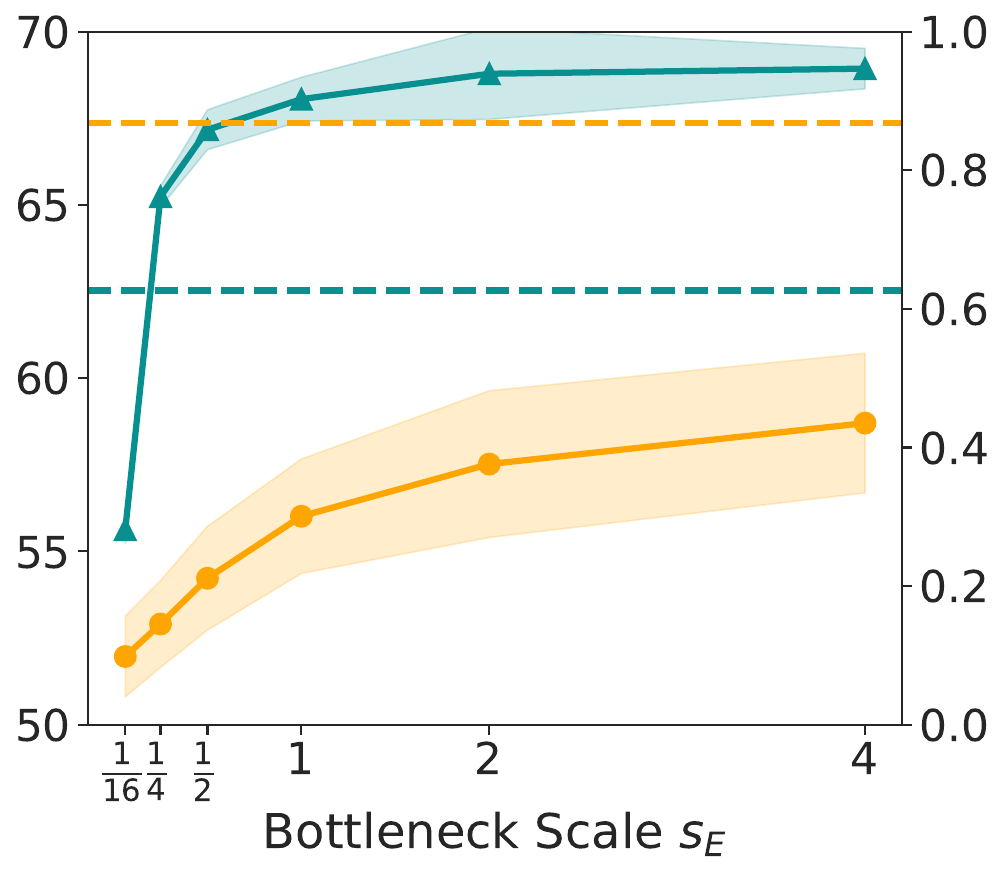}}
\hfil
\subfloat[]{\includegraphics[height=.15\textheight]{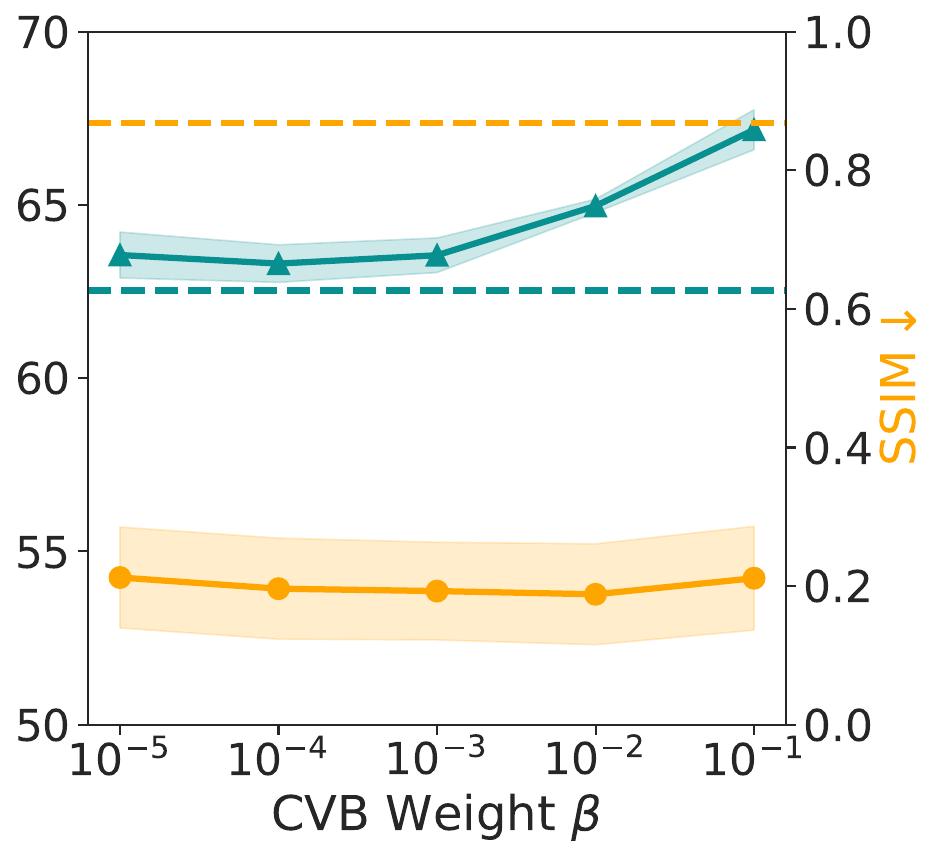}}
\hfil
\caption{
\textbf{Hyperparameter experiments} for a CNN that is protected by our proposed CVB (full line).
The dashed line indicates the baseline for the CNN without protection.
The blue-green lines show model utility measured in terms of accuracy (high is good) and the orange lines show reconstruction quality, \ie, privacy loss measured in terms of SSIM (low is good).
}
\label{fig:ablation}
\end{figure*}

\label{sec:ablation}
We first study the effects of the CVB hyperparameters on privacy and model utility.
The four hyperparameters that affect the CVB are: \textit{1)} the CVB's position $P$ in the network; \textit{2)} the kernel size $k_E$; \textit{3)} the number of kernels $K_E = s_E \cdot K_{E-1}$, where $s_E$ is the relative bottleneck scale with respect to the layer $E-1$ preceding the encoder $E$; and \textit{4)} the weight $\beta$, which regulates the weight of the VB loss on the model loss function.
To save computational resources we limit this hyperparameter study to the CIFAR-10 dataset.
The results of these experiments are displayed in Fig.~\ref{fig:ablation}.
More detailed results can be found in the \textit{Supplementary Material}.

\subsubsection{Position $P$}
First, we vary the position of CVB within the CNN such that $P\in \lbrack 1, 2, 3, (1,2), (1,3), (2,3), (1,2,3) \rbrack$.
The numeral indicates the convolutional layer after which we place the CVB.
With tuples we denote, that we place multiple CVBs at the corresponding positions.
We set $\beta = 0.1$, $k_E=1$ and $s_E=\frac{1}{2}$.
The results are displayed in Fig.~\ref{fig:ablation}(a).

Consistent with the findings in Section~\ref{sec:early_precode} for PRECODE, we find that earlier placement of the CVB correlates with higher privacy.
This is indicated by decreased SSIM from $0.46$ at $P=3$ to $0.13$ at $P=1$.
This corresponds to a drop of ASR from $47.66\%$ to $0\%$.
If the CVB is placed at positions $P=2$ or $P=3$, we can observe a slight increase in model utility compared to the baseline model.
However, at position $P=1$, the accuracy slightly decreases from $62.53\%$ to $62.01\%$.
Adding multiple CVB's to the model does not seem to provide any additional benefits.
Reconstruction quality is mostly impacted by the first placed CVB, \ie, for $P\in \lbrack 1, (1,2), (1,3), (1,2,3) \rbrack$ $\text{SSIM}\approx 0.13$ and for $P\in \lbrack 2, (2,3)\rbrack$ $\text{SSIM}\approx 0.24$.
$P=2$ stands out as the best option with an SSIM of $0.24$ (ASR $0\%$) and the highest accuracy of $65.10\%$.
However, we choose to place the CVB at $P=1$ for all further experiments, because this offers best privacy protection in terms of SSIM.
Furthermore it requires the least amount of additional model parameters.

\subsubsection{Kernel Size $k_E$}
Next, we vary the kernel size of the bottleneck encoder such that $k_E \in \lbrack 1, 3, 5, 7 \rbrack$.
We set $\beta = 0.1$,  $P=1$ and $s_E=\frac{1}{2}$, \ie, $K_{E-1}=16$ and $K_E=8$.
The results are displayed in Fig.~\ref{fig:ablation}(b).

Although the mean reconstruction quality as measured by SSIM correlates with larger $k_E$, the increase in SSIM is negligible and ASR remains constant at $0\%$.
As the number of model parameters increases with larger $k_E$, accuracy also increases from $62.01\%$ for $k_E=1$ to up to $68.23\%$ for $k_E=7$.
We choose to fix $k_E=5$, as it only increases the number of additional model parameters by $9.9\%$ instead of $19.2\%$ for $k_E=7$, while increasing accuracy by $4.64\%$ compared to the baseline CNN.

\subsubsection{Bottleneck Scale $s_E$}
Third, we vary the scale of the bottleneck such that $s_E \in \lbrack \frac{1}{16}, \frac{1}{4}, \frac{1}{2}, 1, 2, 4 \rbrack$.
We set $\beta = 0.1$,  $P=1$ and $k_E=5$.
Note that for $P=1$ the number of channels of the previous layers $K_{E-1}=16$ and hence $K_E \in \lbrack 1, 4, 8, 16, 32, 64 \rbrack$.
The results are displayed in Fig.~\ref{fig:ablation}(c).

As with the kernel size, an increased number of kernels correlates with higher SSIM, accuracy and number of parameters.
We observe an increase in SSIM to up to $0.44$ for $s_E=4$.
This corresponds to an ASR of $26.56\%$.
Accuracy notably increases up to $68.05\%$ for $s_E=1$.
When the bottleneck features become larger than the preceding feature maps, \ie, $s_E>1$, the ensuing improvements become smaller.
If $s_E$ is to small, \ie, $s_E=\frac{1}{16}$, $K_E=1$, accuracy drastically decreases to $55.63\%$
We choose to fix $s_E=\frac{1}{2}$ as it offers the highest gain in accuracy while maintaining a SSIM of $0.21$ (ASR $0\%$).

\subsubsection{CVB Weight $\beta$}
Finally, we vary the weight $\beta$ such that $\beta \in \lbrack 10^{-5}, 10^{-4}, 10^{-3}, 10^{-2}, 10^{-1}\rbrack$
We set $P=1$, $k_E=5$ and $s_E=\frac{1}{2}$.
The results are displayed in Fig.~\ref{fig:ablation}(d).

Larger values for $\beta$ correlate with higher model accuracy.
In all cases model accuracy was increased compared to the baseline of $62.53\%$.
We observe no notable impact on reconstruction quality as indicated by $\text{SSIM}\approx 0.2$.
Except for $\beta = 10^{-5}$ the ASR remains $0\%$.
In the case of $\beta = 10^{-5}$, we were able to reconstruct one out of all $128$ victim images with a $\text{SSIM} \geq 0.5$ ($\text{ASR}=0.78$).
We choose to fix $\beta=10^{-1}$ as it offers the highest gain in accuracy while maintaining an ASR of $0\%$.
In summary, we fix the choice of hyperparameters for the CVB to $P=1$, $k_E=5$, $s_E=\frac{1}{2}$ ($K_E=8$) and $\beta = 0.1$.
For the investigated CNN this requires $9.9\%$ additional model parameters and results in a decrease of SSIM from $0.87$ to $0.21$ (ASR from $96.88\%$ to $0\%$) as well as an increase in accuracy from $62.53\%$ to $67.17\%$ compared to the baseline.

\subsection{Other Attacks}
In the next set of experiments we test the resilience of our proposed CVB against various GI attacks.
Consistent with related work~\cite{wei2020framework, sun2021soteria, li2022auditing, yang2022using, scheliga2023dropout, wei2023securing} we consider three attacks that are commonly used in the literature: iDLG~\cite{zhao2020idlg}, CPL~\cite{wei2020framework} and IG~\cite{geiping2020inverting}.
As described in Section~\ref{sec:sota}, these attacks utilize different optimizers and gradient distance functions.
Again we attack the baseline CNN without defense and the CNN protected by the CVB.
When attacking the CVB protected model, we ignore stochastic gradients as described in Section~\ref{sec:anti_precode}.

\begin{table}
\centering
\caption{
Privacy metrics for a CNN without and with our proposed CVB defense mechanism on MNIST and CIFAR-10. 
The gradients are attacked with different GI attacks. 
Arrows indicate direction of improvement. 
Bold and italic formatting highlight best and worst results respectively.}
\label{tab:sotaattacks}
\resizebox{0.7\linewidth}{!}{%
\begin{tabular}{c|c|c|c|c}
\toprule
 & Defense & Attack & SSIM $\downarrow$ & ASR [\%] $\downarrow$ \\
\midrule
\multirow[c]{7}{*}{\rotatebox{90}{MNIST}} & \multirow[c]{3}{*}{None} & iDLG & {\cellcolor[HTML]{F8864F}} \color[HTML]{F1F1F1} 0.76 ($\pm$0.18) & {\cellcolor[HTML]{CE2827}} \color[HTML]{F1F1F1} 91.41 \\
 &  & CPL & {\cellcolor[HTML]{AFDD70}} \color[HTML]{000000} 0.32 ($\pm$0.17) & {\cellcolor[HTML]{2AA054}} \color[HTML]{F1F1F1} 12.40 \\
 &  & IG & {\cellcolor[HTML]{BD1726}} \color[HTML]{F1F1F1} \itshape 0.95 ($\pm$0.06) & {\cellcolor[HTML]{A50026}} \color[HTML]{F1F1F1} \itshape 100 \\
 \cmidrule{2-5}
 & \multirow[c]{3}{*}{Ours} & iDLG & {\cellcolor[HTML]{006837}} \color[HTML]{F1F1F1} \bfseries 0.00 ($\pm$0.02) & {\cellcolor[HTML]{006837}} \color[HTML]{F1F1F1} \bfseries 0 \\
 &  & CPL & {\cellcolor[HTML]{006837}} \color[HTML]{F1F1F1} \bfseries 0.00 ($\pm$0.02) & {\cellcolor[HTML]{006837}} \color[HTML]{F1F1F1} \bfseries 0 \\
 &  & IG & {\cellcolor[HTML]{A0D669}} \color[HTML]{000000} 0.29 ($\pm$0.09) & {\cellcolor[HTML]{036E3A}} \color[HTML]{F1F1F1} 1.56 \\
\midrule\multirow[c]{7}{*}{\rotatebox{90}{CIFAR-10}} & \multirow[c]{3}{*}{None} & iDLG & {\cellcolor[HTML]{E5F49B}} \color[HTML]{000000} 0.43 ($\pm$0.23) & {\cellcolor[HTML]{F7FCB4}} \color[HTML]{000000} 47.66 \\
 &  & CPL & {\cellcolor[HTML]{AFDD70}} \color[HTML]{000000} 0.32 ($\pm$0.20) & {\cellcolor[HTML]{78C565}} \color[HTML]{000000} 22.66 \\
 &  & IG & {\cellcolor[HTML]{E0422F}} \color[HTML]{F1F1F1} \itshape 0.87 ($\pm$0.11) & {\cellcolor[HTML]{B30D26}} \color[HTML]{F1F1F1} \itshape 96.88 \\
 \cmidrule{2-5}
 & \multirow[c]{3}{*}{Ours} & iDLG & {\cellcolor[HTML]{026C39}} \color[HTML]{F1F1F1} \bfseries 0.01 ($\pm$0.01) & {\cellcolor[HTML]{006837}} \color[HTML]{F1F1F1} \bfseries 0 \\
 &  & CPL & {\cellcolor[HTML]{07753E}} \color[HTML]{F1F1F1} 0.03 ($\pm$0.02) & {\cellcolor[HTML]{006837}} \color[HTML]{F1F1F1} \bfseries 0 \\
 &  & IG & {\cellcolor[HTML]{6BBF64}} \color[HTML]{000000} 0.21 ($\pm$0.08) & {\cellcolor[HTML]{006837}} \color[HTML]{F1F1F1} \bfseries 0 \\
\bottomrule
\end{tabular}
}
\end{table}

Tab.~\ref{tab:sotaattacks} shows that IG achieves the best reconstruction qualities as indicated by larger SSIM and ASR.
In all cases the gradients of the baseline model without defense are vulnerable to GI attacks.
The CVB on the other hand protects privacy as indicated by an ASR of $0\%$.
Only for the MNIST dataset and the IG attacker, we were able to reconstruct $2$ out of all $128$ victim images with a $\text{SSIM} \geq 0.5$ ($\text{ASR}=1.56$).
As the CVB reduces the amount of gradient information which is usable for the attack optimization, it protects from various iterative GI attacks.

\subsection{Other Defenses}
Next, we compare our proposed CVB to other common gradient perturbation techniques and PRECODE.
For gradient perturbation we consider noisy gradients and gradient compression, each with different perturbation levels.
Gradient quantization is not considered due to the comparatively large negative impact on model utility~\cite{zhu2019deep}.

For noisy gradients, we utilize Opacus'~\cite{yousefpour2021opacus} implementation of differentially private SGD (DPSGD)~\cite{abadi2016deep}, denoted as DP.
The DPSGD algorithm first calculates the per-sample gradient for each sample in an input batch and clips it with clipping threshold $C$.
Then Gaussian distributed noise with zero mean and standard deviation $C \cdot\sigma$ is added.
To find a practical clipping threshold $C$, we follow~\cite{ponomareva2023dp} and perform some preliminary experiments to find the smallest $C$ that only marginally harms model utility.
We search for $C \in \lbrace 1, 10, 20, 30, 50, 100 \rbrace$ and eventually fix $C=20$
We increase noise levels by setting the noise multiplier $\sigma \in \lbrack 10^{-3}, 10^{-2}, 10^{-1}, 1 \rbrack$.
For GC the gradient values with smallest magnitudes are pruned to zero for every gradient layer, as described in~\cite{huang2021evaluating}.
The amount of pruned values is determined by the pruning ratio $p$, which we set to $p \in [0.9, 0.99]$.
For PRECODE we insert the VB at position $P=1$ to offer most protection.
We fix $\beta=0.01$ and vary the size of the bottleneck such that $K \in \lbrack 32, 64, 128, 256, 512 \rbrack$.
We apply our proposed CVB with the parameters determined in Section~\ref{sec:ablation}, \ie, $P=1$, $k_E=5$, $s_E=\frac{1}{2}$ and $\beta = 0.1$.
Whenever we attack the gradients of a model defended with PRECODE or our CVB, we utilize our proposed "Ignore" attack (\cf Section~\ref{sec:anti_precode}).
The results for these experiments are displayed in Tab.~\ref{tab:sotadefenses}.

Consistent with related work, we observe the typical utility-privacy trade-off when perturbation-based defenses, \ie, DP and GC are applied.
In order to reduce the ASR to $0\%$, DP requires a noise multiplier of $\sigma=10^{-2}$ and GC a pruning rate of $p=0.99$.
This in turn decreases the model utility in terms of accuracy by $0.23\%/1.76\%$ for DP and $76.02\%/51.04\%$ for GC for the MNIST/CIFAR-10 datasets respectively.
For PRECODE we observe a consistent decrease of reconstruction quality in terms of SSIM and ASR for smaller bottleneck sizes $K$.
However, for all investigated cases we were able to reconstruct at least one sample with a $\text{SSIM}>0.5$.
If $K$ is to large, \eg $K=512$, almost all samples can be reconstructed.
This is reflected by an ASR of $100\%$ for MNIST and $97.66$ for CIFAR-10.
The larger $K$, the more gradient information is available for the attack optimization, because the gradients of the encoding bottleneck layer $E$ can be attacked.
Only layers that succeed this layer are protected by the stochastic effects of PRECODE.
Too small or too large bottleneck sizes result in a reduced accuracy.
The best accuracies of $98.55\%$ and $61.60\%$ were achieved with $K=32$ and $K=64$ for MNIST and CIFAR-10 respectively. 
However, for all $K$ accuracy decreases compared to the baseline.
In comparison, our proposed CVB reduces ASR to $0.78\%$ and $0\%$ while also improving accuracy by $0.05\%$ and $4.64\%$ for MNIST and CIFAR-10 respectively.

\begin{table}
\centering
\caption{
Model utility and privacy metrics for a CNN without and with defense trained on MNIST and CIFAR-10. 
The gradients are attacked for different defense mechanisms.
Parameters indicate clipping threshold and noise multiplier $(C, \sigma)$ for DP; pruning ratio $p$ for GC; bottleneck size $K$ for PRECODE, as well as kernel size and bottleneck scale $(k_E, s_E)$ for our proposed CVB.
}
\label{tab:sotadefenses}
\resizebox{0.99\linewidth}{!}{%
\begin{tabular}{c|c|c|c|c|c}
\toprule
 & Defense & Parameters & SSIM $\downarrow$ & ASR [\%] $\downarrow$ & Accuracy [\%] $\uparrow$ \\
\midrule
\multirow[c]{15}{*}{\rotatebox{90}{MNIST}} & None & - & {\cellcolor[HTML]{BD1726}} \color[HTML]{F1F1F1} 0.95 ($\pm$0.06) & {\cellcolor[HTML]{A50026}} \color[HTML]{F1F1F1} \itshape 100 & {\cellcolor[HTML]{006837}} \color[HTML]{F1F1F1} 99.10 ($\pm$0.04) \\
\cmidrule{2-6}
 & \multirow[c]{4}{*}{DP} & $(20, 1)$ & {\cellcolor[HTML]{006837}} \color[HTML]{F1F1F1} \bfseries 0.00 ($\pm$0.03) & {\cellcolor[HTML]{006837}} \color[HTML]{F1F1F1} \bfseries 0 & {\cellcolor[HTML]{006837}} \color[HTML]{F1F1F1} 94.73 ($\pm$0.12) \\
 &  & $(20, 10^{-1})$ & {\cellcolor[HTML]{026C39}} \color[HTML]{F1F1F1} 0.01 ($\pm$0.03) & {\cellcolor[HTML]{006837}} \color[HTML]{F1F1F1} \bfseries 0 & {\cellcolor[HTML]{006837}} \color[HTML]{F1F1F1} 98.17 ($\pm$0.07) \\
 &  & $(20, 10^{-2})$ & {\cellcolor[HTML]{148E4B}} \color[HTML]{F1F1F1} 0.08 ($\pm$0.04) & {\cellcolor[HTML]{006837}} \color[HTML]{F1F1F1} \bfseries 0 & {\cellcolor[HTML]{006837}} \color[HTML]{F1F1F1} 98.87 ($\pm$0.08) \\
 &  & $(20, 10^{-3})$ & {\cellcolor[HTML]{FECC7B}} \color[HTML]{000000} 0.64 ($\pm$0.11) & {\cellcolor[HTML]{E44C34}} \color[HTML]{F1F1F1} 85.16 & {\cellcolor[HTML]{006837}} \color[HTML]{F1F1F1} 98.97 ($\pm$0.06) \\
 \cmidrule{2-6}
 & \multirow[c]{2}{*}{GC} & $0.9$ & {\cellcolor[HTML]{FED07E}} \color[HTML]{000000} 0.63 ($\pm$0.09) & {\cellcolor[HTML]{D83128}} \color[HTML]{F1F1F1} 89.84 & {\cellcolor[HTML]{006837}} \color[HTML]{F1F1F1} 98.31 ($\pm$0.07) \\
 &  & $0.99$ & {\cellcolor[HTML]{7FC866}} \color[HTML]{000000} 0.24 ($\pm$0.05) & {\cellcolor[HTML]{006837}} \color[HTML]{F1F1F1} \bfseries 0 & {\cellcolor[HTML]{A50026}} \color[HTML]{F1F1F1} \itshape 23.08 ($\pm$9.39) \\
 \cmidrule{2-6}
 & \multirow[c]{5}{*}{PRECODE} & 32 & {\cellcolor[HTML]{EFF8AA}} \color[HTML]{000000} 0.46 ($\pm$0.09) & {\cellcolor[HTML]{B7E075}} \color[HTML]{000000} 33.59 & {\cellcolor[HTML]{006837}} \color[HTML]{F1F1F1} 98.55 ($\pm$0.09) \\
 &  & 64 & {\cellcolor[HTML]{FFF8B4}} \color[HTML]{000000} 0.52 ($\pm$0.10) & {\cellcolor[HTML]{FEE18D}} \color[HTML]{000000} 59.38 & {\cellcolor[HTML]{006837}} \color[HTML]{F1F1F1} 98.42 ($\pm$0.06) \\
 &  & 128 & {\cellcolor[HTML]{FEE28F}} \color[HTML]{000000} 0.59 ($\pm$0.10) & {\cellcolor[HTML]{EC5C3B}} \color[HTML]{F1F1F1} 82.81 & {\cellcolor[HTML]{006837}} \color[HTML]{F1F1F1} 98.49 ($\pm$0.10) \\
 &  & 256 & {\cellcolor[HTML]{FDB768}} \color[HTML]{000000} 0.68 ($\pm$0.10) & {\cellcolor[HTML]{D42D27}} \color[HTML]{F1F1F1} 90.62 & {\cellcolor[HTML]{006837}} \color[HTML]{F1F1F1} 98.37 ($\pm$0.18) \\
 &  & 512 & {\cellcolor[HTML]{F67A49}} \color[HTML]{F1F1F1} 0.78 ($\pm$0.09) & {\cellcolor[HTML]{A50026}} \color[HTML]{F1F1F1} \itshape 100 & {\cellcolor[HTML]{006837}} \color[HTML]{F1F1F1} 98.22 ($\pm$0.07) \\
 \cmidrule{2-6}
 & \textbf{Ours} & $(5,\frac{1}{2})$ & {\cellcolor[HTML]{A0D669}} \color[HTML]{000000} 0.29 ($\pm$0.09) & {\cellcolor[HTML]{016A38}} \color[HTML]{F1F1F1} 0.78 & {\cellcolor[HTML]{006837}} \color[HTML]{F1F1F1} \bfseries 99.15 ($\pm$0.05) \\
\midrule\multirow[c]{15}{*}{\rotatebox{90}{CIFAR-10}} & None & - & {\cellcolor[HTML]{E0422F}} \color[HTML]{F1F1F1} 0.87 ($\pm$0.11) & {\cellcolor[HTML]{B30D26}} \color[HTML]{F1F1F1} 96.88 & {\cellcolor[HTML]{CBE982}} \color[HTML]{000000} 62.53 ($\pm$0.16) \\
 \cmidrule{2-6}
 & \multirow[c]{4}{*}{DP} & $(20, 1)$ & {\cellcolor[HTML]{026C39}} \color[HTML]{F1F1F1} \bfseries 0.01 ($\pm$0.02) & {\cellcolor[HTML]{006837}} \color[HTML]{F1F1F1} \bfseries 0 & {\cellcolor[HTML]{A50026}} \color[HTML]{F1F1F1} 49.87 ($\pm$1.19) \\
 &  & $(20, 10^{-1})$ & {\cellcolor[HTML]{026C39}} \color[HTML]{F1F1F1} \bfseries 0.01 ($\pm$0.02) & {\cellcolor[HTML]{006837}} \color[HTML]{F1F1F1} \bfseries 0 & {\cellcolor[HTML]{FFF0A6}} \color[HTML]{000000} 59.02 ($\pm$0.91) \\
 &  & $(20, 10^{-2})$ & {\cellcolor[HTML]{07753E}} \color[HTML]{F1F1F1} 0.03 ($\pm$0.02) & {\cellcolor[HTML]{006837}} \color[HTML]{F1F1F1} \bfseries 0 & {\cellcolor[HTML]{F1F9AC}} \color[HTML]{000000} 60.77 ($\pm$0.41) \\
 &  & $(20, 10^{-3})$ & {\cellcolor[HTML]{ABDB6D}} \color[HTML]{000000} 0.31 ($\pm$0.11) & {\cellcolor[HTML]{0E8245}} \color[HTML]{F1F1F1} 5.47 & {\cellcolor[HTML]{F1F9AC}} \color[HTML]{000000} 60.73 ($\pm$0.25) \\
 \cmidrule{2-6}
 & \multirow[c]{2}{*}{GC} & $0.9$ & {\cellcolor[HTML]{F7FCB4}} \color[HTML]{000000} 0.48 ($\pm$0.08) & {\cellcolor[HTML]{F4FAB0}} \color[HTML]{000000} 46.88 & {\cellcolor[HTML]{F67A49}} \color[HTML]{F1F1F1} 54.38 ($\pm$1.73) \\
 &  & $0.99$ & {\cellcolor[HTML]{3FAA59}} \color[HTML]{F1F1F1} 0.15 ($\pm$0.04) & {\cellcolor[HTML]{006837}} \color[HTML]{F1F1F1} \bfseries 0 & {\cellcolor[HTML]{A50026}} \color[HTML]{F1F1F1} \itshape 11.49 ($\pm$1.29) \\
 \cmidrule{2-6}
 & \multirow[c]{5}{*}{PRECODE} & 32 & {\cellcolor[HTML]{AFDD70}} \color[HTML]{000000} 0.32 ($\pm$0.09) & {\cellcolor[HTML]{036E3A}} \color[HTML]{F1F1F1} 1.56 & {\cellcolor[HTML]{E2F397}} \color[HTML]{000000} 61.49 ($\pm$0.51) \\
 &  & 64 & {\cellcolor[HTML]{BFE47A}} \color[HTML]{000000} 0.35 ($\pm$0.10) & {\cellcolor[HTML]{138C4A}} \color[HTML]{F1F1F1} 7.81 & {\cellcolor[HTML]{E0F295}} \color[HTML]{000000} 61.60 ($\pm$0.38) \\
 &  & 128 & {\cellcolor[HTML]{DCF08F}} \color[HTML]{000000} 0.41 ($\pm$0.11) & {\cellcolor[HTML]{60BA62}} \color[HTML]{F1F1F1} 19.53 & {\cellcolor[HTML]{F2FAAE}} \color[HTML]{000000} 60.65 ($\pm$0.70) \\
 &  & 256 & {\cellcolor[HTML]{FFF6B0}} \color[HTML]{000000} 0.53 ($\pm$0.12) & {\cellcolor[HTML]{FED683}} \color[HTML]{000000} 61.72 & {\cellcolor[HTML]{ECF7A6}} \color[HTML]{000000} 60.97 ($\pm$0.55) \\
 &  & 512 & {\cellcolor[HTML]{FBA05B}} \color[HTML]{000000} 0.72 ($\pm$0.09) & {\cellcolor[HTML]{AF0926}} \color[HTML]{F1F1F1} \itshape 97.66 & {\cellcolor[HTML]{FFFEBE}} \color[HTML]{000000} 59.95 ($\pm$0.11) \\
 \cmidrule{2-6}
 & \textbf{Ours} & $(5,\frac{1}{2})$ & {\cellcolor[HTML]{6BBF64}} \color[HTML]{000000} 0.21 ($\pm$0.08) & {\cellcolor[HTML]{006837}} \color[HTML]{F1F1F1} \bfseries 0 & {\cellcolor[HTML]{39A758}} \color[HTML]{F1F1F1} \bfseries 67.17 ($\pm$0.58) \\
\bottomrule
\end{tabular}
}
\end{table}

\subsection{Other Model Architectures}
\begin{figure}
    \centering
    \includegraphics[width=.99\linewidth]{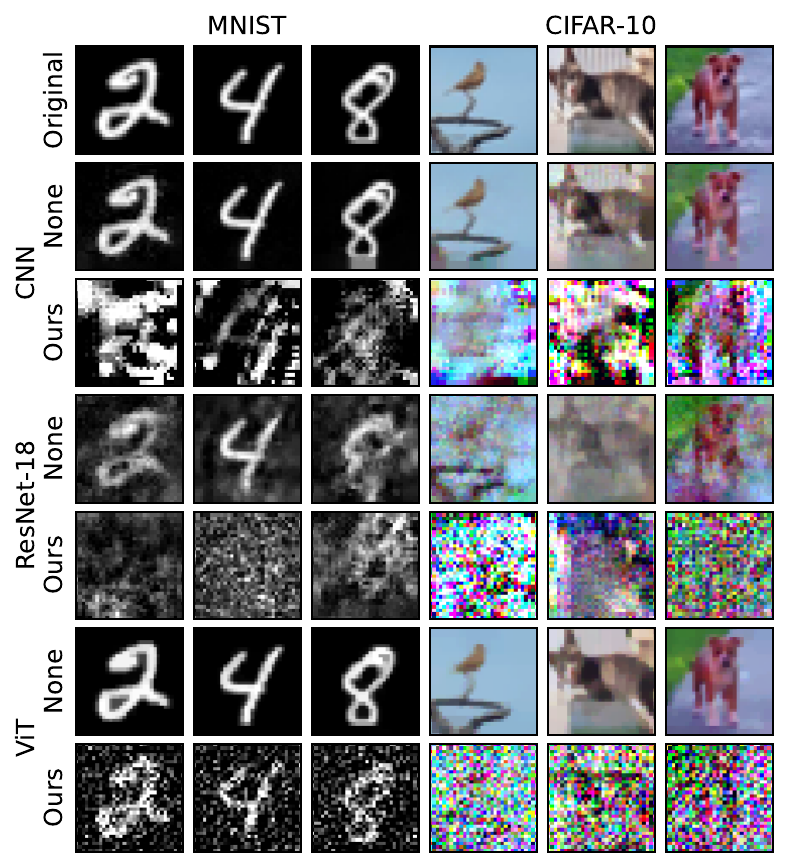}
     \caption{\textbf{Example reconstructions} for different model architectures without and with our proposed defense on the MNIST and CIFAR-10 datasets.}
     \label{fig:model_rec}
\end{figure}

Next, we apply the CVB to different model architectures.
For model architectures we consider a ResNet-18~\cite{he2016deep} and small Vision Transformer (ViT)~\cite{dosovitskiy2020image}.
In the ResNet-18 we place a CVB with kernel size $k_E=5$ and $s_E=\frac{1}{2}$ ($K_E=32$) channels after the first bottleneck block.
We did a parameter search for $\beta \in \lbrack 10^{-1}, 10^{-2}, 10^{-3}, 10^{-4}, 5\cdot 10^{-5}, 10^{-5} \rbrack$ on the CIFAR-10 dataset and found $\beta=5\cdot 10^{-5}$ to result in the best model accuracy.
In the ViT we place a CVB with kernel size $k_E=3$ and $s_E=\frac{1}{2}$ ($K_E=32$) channels after the first transformer block.
Larger kernel sizes resulted in unstable training.
The corresponding parameter search for $\beta$ resulted in a choice of $\beta=0.1$, because it offered the best model accuracy.
To apply the convolutional operations of the CVB to the transformer we reshape features from $(\mathrm{P} \times \mathrm{F})$ to $(\mathrm{P} \times \mathrm{H} \times \mathrm{W})$ to be two-dimensional, where the patch dimension $\mathrm{P}$ is handled as channels.
Note, that this requires the hidden dimension $\mathrm{F}$ of the transformer to be $\mathrm{F}=\mathrm{H} \cdot \mathrm{W}$.
Hence, we choose the following architecture parameters for the ViT: embedding patch size $4$, $4$ transformer blocks with a hidden size $\mathrm{F}=256$ ($\mathrm{H}=\mathrm{W}=16$), $16$ attention heads and GELU~\cite{hendrycks2016gaussian} activation.
Tab.~\ref{tab:model} displays the results of these experiments.
Fig.~\ref{fig:model_rec} displays some example reconstructions.

\begin{table}
\centering
\caption{
Privacy and model utility metrics for a CNN, ResNet-18 and ViT without and with our proposed CVB defense trained on MNIST and CIFAR-10. 
}
\label{tab:model}
\resizebox{0.99\linewidth}{!}{%
\begin{tabular}{c|c|c|c|c|c}
\toprule
 & Model & Defense & SSIM $\downarrow$ & ASR [\%] $\downarrow$ & Accuracy [\%] $\uparrow$ \\
\midrule
\multirow[c]{7}{*}{\rotatebox{90}{MNIST}} & \multirow[c]{2}{*}{CNN} & None & {\cellcolor[HTML]{BD1726}} \color[HTML]{F1F1F1} \itshape 0.95 ($\pm$0.06) & {\cellcolor[HTML]{A50026}} \color[HTML]{F1F1F1} \itshape 100 & {\cellcolor[HTML]{006837}} \color[HTML]{F1F1F1} \itshape 99.10 ($\pm$0.04) \\
 &   & Ours & {\cellcolor[HTML]{A0D669}} \color[HTML]{000000} \bfseries 0.29 ($\pm$0.09) & {\cellcolor[HTML]{016A38}} \color[HTML]{F1F1F1} \bfseries 0.78 & {\cellcolor[HTML]{006837}} \color[HTML]{F1F1F1} \bfseries 99.15 ($\pm$0.05) \\
\cmidrule{2-6}
 & \multirow[c]{2}{*}{ResNet-18} & None & {\cellcolor[HTML]{FFFCBA}} \color[HTML]{000000} \itshape 0.51 ($\pm$0.12) & {\cellcolor[HTML]{FFF2AA}} \color[HTML]{000000} \itshape 53.91 & {\cellcolor[HTML]{006837}} \color[HTML]{F1F1F1} \itshape 99.39 ($\pm$0.09) \\
 &   & Ours & {\cellcolor[HTML]{05713C}} \color[HTML]{F1F1F1} \bfseries 0.02 ($\pm$0.04) & {\cellcolor[HTML]{006837}} \color[HTML]{F1F1F1} \bfseries 0 & {\cellcolor[HTML]{006837}} \color[HTML]{F1F1F1} \bfseries 99.46 ($\pm$0.04) \\
\cmidrule{2-6}
 & \multirow[c]{2}{*}{ViT} & None & {\cellcolor[HTML]{A90426}} \color[HTML]{F1F1F1} \itshape 0.99 ($\pm$0.00) & {\cellcolor[HTML]{A50026}} \color[HTML]{F1F1F1} \itshape 100 & {\cellcolor[HTML]{006837}} \color[HTML]{F1F1F1} \itshape 98.70 ($\pm$0.14) \\
 &   & Ours & {\cellcolor[HTML]{C9E881}} \color[HTML]{000000} \bfseries 0.37 ($\pm$0.08) & {\cellcolor[HTML]{118848}} \color[HTML]{F1F1F1} \bfseries 7.03 & {\cellcolor[HTML]{006837}} \color[HTML]{F1F1F1} \bfseries 98.96 ($\pm$0.04) \\
\midrule\multirow[c]{7}{*}{\rotatebox{90}{CIFAR-10}} & \multirow[c]{2}{*}{CNN} & None & {\cellcolor[HTML]{E0422F}} \color[HTML]{F1F1F1} \itshape 0.87 ($\pm$0.11) & {\cellcolor[HTML]{B30D26}} \color[HTML]{F1F1F1} \itshape 96.88 & {\cellcolor[HTML]{CBE982}} \color[HTML]{000000} \itshape 62.53 ($\pm$0.16) \\
 &   & Ours & {\cellcolor[HTML]{6BBF64}} \color[HTML]{000000} \bfseries 0.21 ($\pm$0.08) & {\cellcolor[HTML]{006837}} \color[HTML]{F1F1F1} \bfseries 0 & {\cellcolor[HTML]{39A758}} \color[HTML]{F1F1F1} \bfseries 67.17 ($\pm$0.58) \\
\cmidrule{2-6}
 & \multirow[c]{2}{*}{ResNet-18} & None & {\cellcolor[HTML]{FFF0A6}} \color[HTML]{000000} \itshape 0.55 ($\pm$0.11) & {\cellcolor[HTML]{FDB365}} \color[HTML]{000000} \itshape 68.75 & {\cellcolor[HTML]{006837}} \color[HTML]{F1F1F1} \itshape 73.18 ($\pm$0.37) \\
 &   & Ours & {\cellcolor[HTML]{17934E}} \color[HTML]{F1F1F1} \bfseries 0.09 ($\pm$0.07) & {\cellcolor[HTML]{006837}} \color[HTML]{F1F1F1} \bfseries 0 & {\cellcolor[HTML]{006837}} \color[HTML]{F1F1F1} \bfseries 74.17 ($\pm$0.20) \\
\cmidrule{2-6}
 & \multirow[c]{2}{*}{ViT} & None & {\cellcolor[HTML]{D62F27}} \color[HTML]{F1F1F1} \itshape 0.90 ($\pm$0.05) & {\cellcolor[HTML]{A50026}} \color[HTML]{F1F1F1} \itshape 100 & {\cellcolor[HTML]{B5DF74}} \color[HTML]{000000} \itshape 63.39 ($\pm$0.50) \\
 &   & Ours & {\cellcolor[HTML]{45AD5B}} \color[HTML]{F1F1F1} \bfseries 0.16 ($\pm$0.06) & {\cellcolor[HTML]{006837}} \color[HTML]{F1F1F1} \bfseries 0 & {\cellcolor[HTML]{B3DF72}} \color[HTML]{000000} \bfseries 63.50 ($\pm$0.34) \\


\bottomrule
\end{tabular}
}
\end{table}

All results confirm the previous findings made for the CNN.
Our proposed CVB significantly decreases gradient leakage for all architectures.
Attack success for the unprotected CNN and ViT is close to $100\%$ for all datasets.
For the unprotected ResNet-18 $53.91\%$ and $68.75\%$ of the data can be successfully reconstructed for the MNIST and CIFAR-10 datasets respectively.
We attribute this inherent baseline protection against GI attacks to the size of the network and therefore increased complexity of the attack optimization.
If we add our proposed CVB to the model, the ASR drops to $0\%$ for most dataset-architecture combinations.
The noticeable drops in SSIM reflect this decrease in reconstruction quality accordingly.
Only for the CNN and ViT with less complex datasets, \ie, MNIST, there are a few successful reconstructions.
We conclude that increased diversity of training data inherently increases protection of most architectures.
Compared to the ViT, we observe lower reconstruction quality in terms of SSIM for architectures based on convolution layers, \ie, CNN and ResNet-18.
CNNs can inherently offer some protection against gradient leakage due to local connectivity and parameter sharing of sliding convolutions.
The ViT also uses shared parameters that are locally connected to each image patch.
However, unlike CNNs, these patches do not overlap, which enables improved reconstruction of inputs for the ViT.
Additionally, the use or CVB results in an increase of model utility for all datasets and architectures.
Similar regularizing effects when extending architectures with variational modeling were reported by~\cite{alemi2016deep, hofmann2021synaptic}.
The most significant increase in accuracy of $4.64\%$ can be observed for the CNN trained on the CIFAR-10 dataset.

\begin{figure*}[!t]
\centering
\subfloat[]{\includegraphics[height=.2\textheight]{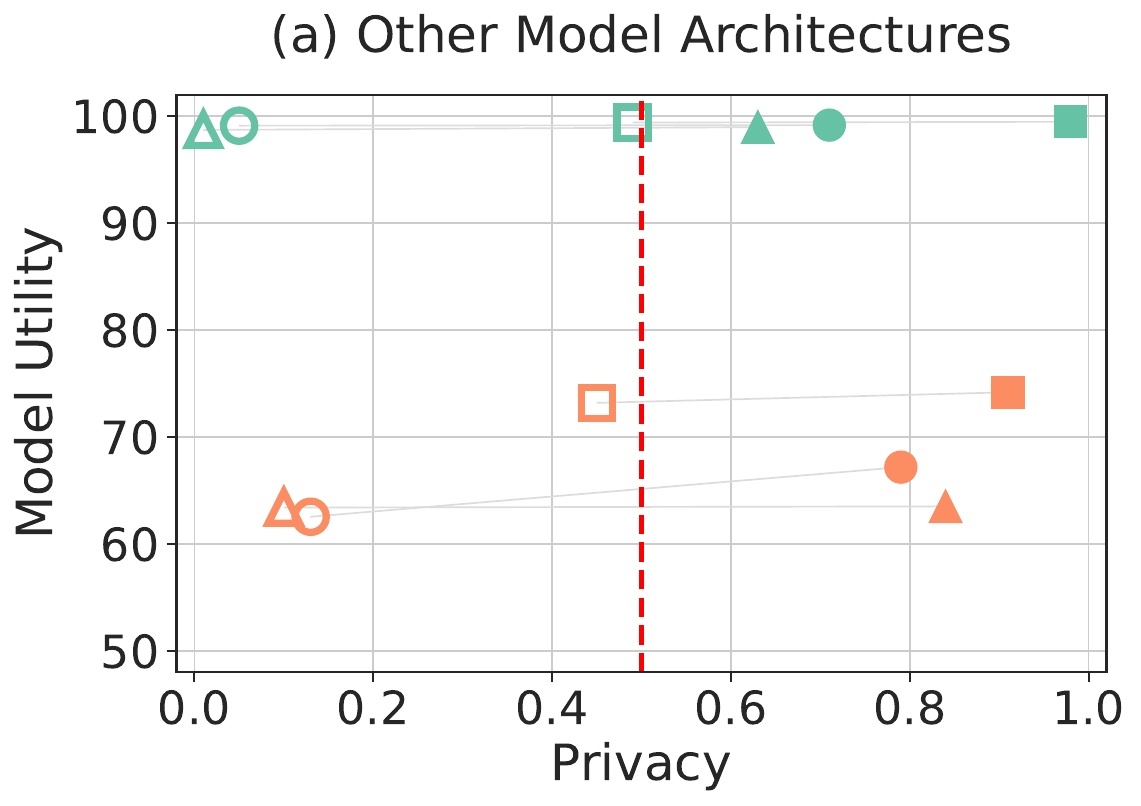}}
\hfil
\subfloat[]{\includegraphics[height=.2\textheight]{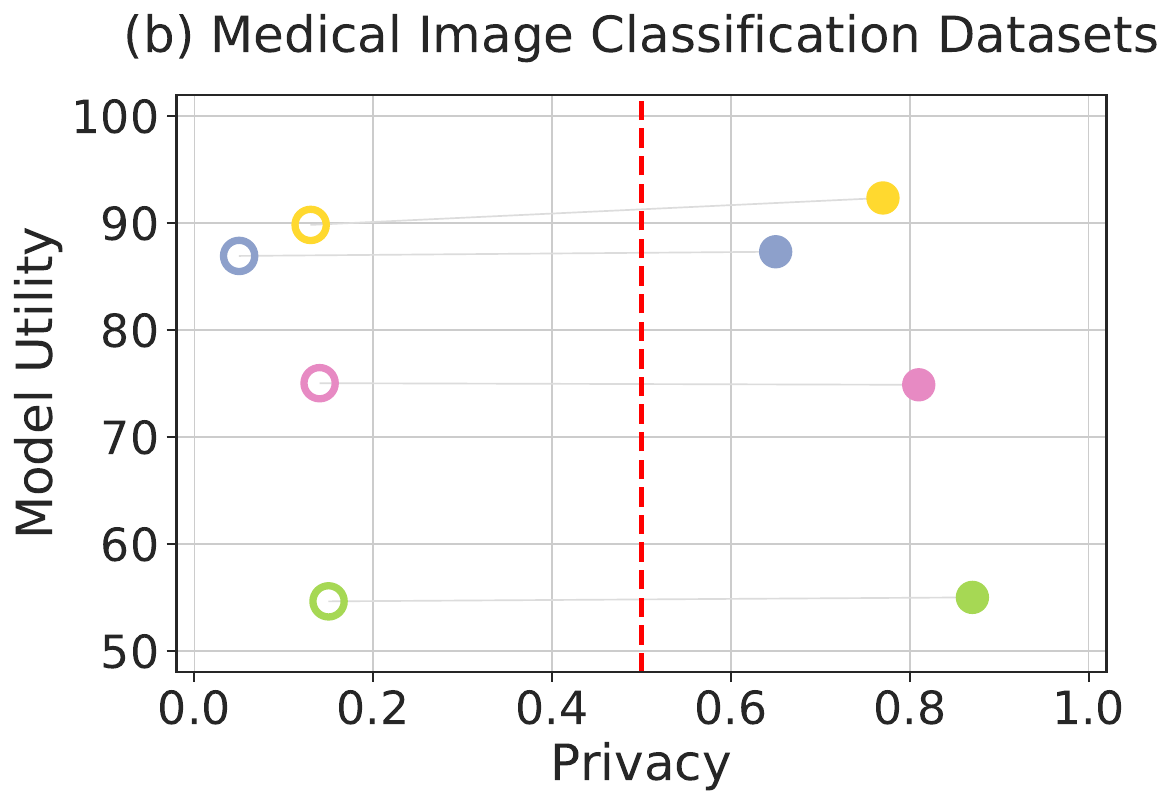}}
\hfil
\subfloat[]{\includegraphics[height=.2\textheight]{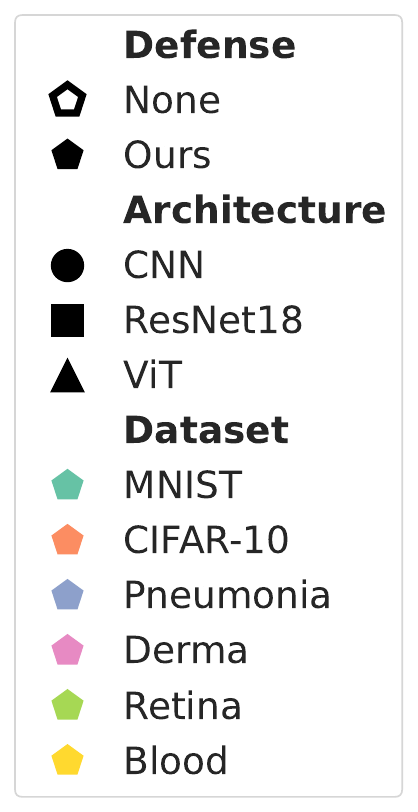}}
\hfil
\caption{\textbf{Model utility -- Privacy trade-off} for (a) different model architectures on the MNIST and CIFAR-10; and (b) a CNN on different MedMNIST datasets.
    Model utility and privacy are displayed in terms of test accuracy and $1-$SSIM. 
    Color indicates the dataset.
    Model architectures are distinguished by marker symbols. 
    Empty symbols display models without defense whereas filled symbols display models extended by our proposed CVB.
    The red line indicates the threshold for attack success.
}
\label{fig:takeaways}
\end{figure*}

To visually summarize the trade-off between model utility and privacy for all considered datasets and models, we aggregated the results into Fig.~\ref{fig:takeaways}(a).
Model utility and privacy are represented as test accuracy and $1-$SSIM respectively.
An optimal model would have perfect model utility and privacy, locating itself in the top right corner of the plot.
Upon extending models with the CVB, a shift towards increased utility and privacy can be observed for all model architectures.
The red line in Fig.~\ref{fig:takeaways} denotes the privacy threshold.
Models located to the right of the threshold can be considered privacy preserving, as GI attacks result in reconstruction of training data with no more than $0.5$ SSIM on average.

\subsection{Use Case: Medical Image Classification}

\begin{table}
\centering
\caption{
Privacy and model utility metrics for a CNN without and with our proposed CVB defense mechanism trained on four MedMNIST datasets. 
}
\label{tab:datasets}
\resizebox{0.99\linewidth}{!}{%
\begin{tabular}{c|c|c|c|c|c|c|c|c|c}
\toprule
 Dataset & Defense & SSIM $\downarrow$ & ASR [\%] $\downarrow$ & Accuracy [\%] $\uparrow$ \\
\midrule
\multirow[c]{2}{*}{Pneumonia} & None & {\cellcolor[HTML]{BD1726}} \color[HTML]{F1F1F1} \itshape 0.95 ($\pm$0.04) & {\cellcolor[HTML]{A50026}} \color[HTML]{F1F1F1} \itshape 100 & {\cellcolor[HTML]{006837}} \color[HTML]{F1F1F1} \itshape 86.92 ($\pm$0.21) \\
 & Ours & {\cellcolor[HTML]{BFE47A}} \color[HTML]{000000} \bfseries 0.35 ($\pm$0.11) & {\cellcolor[HTML]{138C4A}} \color[HTML]{F1F1F1} \bfseries 7.81 & {\cellcolor[HTML]{006837}} \color[HTML]{F1F1F1} \bfseries 87.31 ($\pm$1.22) \\
\midrule\multirow[c]{2}{*}{Derma} & None & {\cellcolor[HTML]{E24731}} \color[HTML]{F1F1F1} \itshape 0.86 ($\pm$0.13) & {\cellcolor[HTML]{B30D26}} \color[HTML]{F1F1F1} \itshape 96.88 & {\cellcolor[HTML]{006837}} \color[HTML]{F1F1F1} \bfseries 75.02 ($\pm$0.40) \\
 & Ours & {\cellcolor[HTML]{5DB961}} \color[HTML]{F1F1F1} \bfseries 0.19 ($\pm$0.10) & {\cellcolor[HTML]{006837}} \color[HTML]{F1F1F1} \bfseries 0 & {\cellcolor[HTML]{006837}} \color[HTML]{F1F1F1} \itshape 74.87 ($\pm$0.69) \\
\midrule\multirow[c]{2}{*}{Retina} & None & {\cellcolor[HTML]{E54E35}} \color[HTML]{F1F1F1} \itshape 0.85 ($\pm$0.14) & {\cellcolor[HTML]{B91326}} \color[HTML]{F1F1F1} \itshape 96.09 & {\cellcolor[HTML]{006837}} \color[HTML]{F1F1F1} \itshape 54.62 ($\pm$1.97) \\
 & Ours & {\cellcolor[HTML]{30A356}} \color[HTML]{F1F1F1} \bfseries 0.13 ($\pm$0.06) & {\cellcolor[HTML]{006837}} \color[HTML]{F1F1F1} \bfseries 0 & {\cellcolor[HTML]{006837}} \color[HTML]{F1F1F1} \bfseries 54.99 ($\pm$1.00) \\
\midrule\multirow[c]{2}{*}{Blood} & None & {\cellcolor[HTML]{E0422F}} \color[HTML]{F1F1F1} \itshape 0.87 ($\pm$0.11) & {\cellcolor[HTML]{B30D26}} \color[HTML]{F1F1F1} \itshape 96.88 & {\cellcolor[HTML]{006837}} \color[HTML]{F1F1F1} \itshape 89.82 ($\pm$0.08) \\
 & Ours & {\cellcolor[HTML]{78C565}} \color[HTML]{000000} \bfseries 0.23 ($\pm$0.08) & {\cellcolor[HTML]{006837}} \color[HTML]{F1F1F1} \bfseries 0 & {\cellcolor[HTML]{006837}} \color[HTML]{F1F1F1} \bfseries 92.34 ($\pm$0.13) \\
\bottomrule
\end{tabular}
}
\end{table}

\begin{figure}
    \centering
    \includegraphics[width=.99\linewidth]{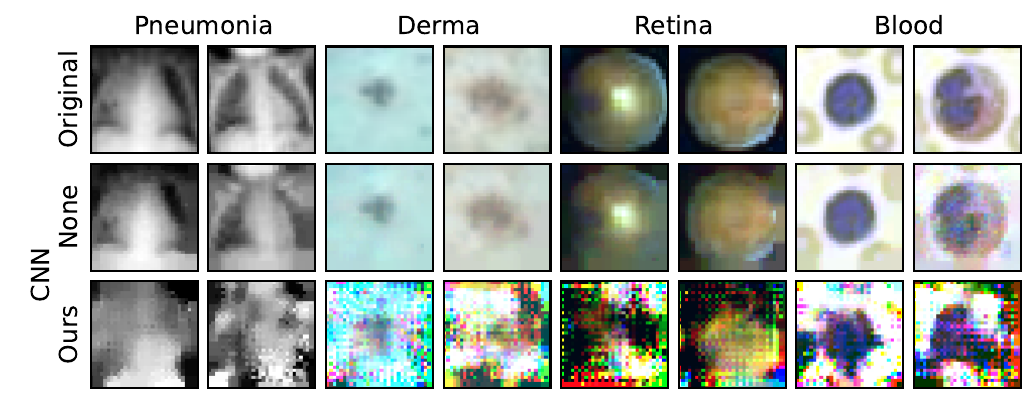}
     \caption{\textbf{Example reconstructions} for a CNN without and with our proposed defense on different MedMNIST datasets.}
     \label{fig:datasets_rec}
\end{figure}

Finally, we present a practical use case by evaluating our approach on a set of medical image classification datasets.
Specifically, we consider the MedMNIST-Pneumonia, MedMNIST-Derma, MedMNIST-Retina and MedMNIST-Blood datasets~\cite{yang2023medmnist}.
MedMNIST-Pneumonia consists of $5,856$ pediatric chest X-ray images with the task to discriminate between pneumonia and normal.
MedMNIST-Derma consists of $10,015$ dermatoscopic images with the task to classify seven different diseases.
MedMNIST-Retina consists of $1,600$ retina fundus images with the task to grade five levels of severity of diabetic retinopathy.
MedMNIST-Blood consists of $17,092$ images of blood-cells that were captured from individuals at the moment of blood collection with the task to classify eight different types of blood-cells.
We apply the data distribution protocols as described in Section~\ref{sec:design}.
For each dataset, we perform a brief parameter search to determine valid values of $k_E \in \lbrack 3, 5 \rbrack$ and $\beta \in \lbrack 10^{-1}, 10^{-2}, 10^{-3} \rbrack$ when using our CVB.
We fixed $(k_E, \beta)$ to $(5, 10^{-3})$ for Peumonia, $(3, 10^{-1})$ for Derma, $(3, 10^{-3})$ for Retina and $(5, 10^{-1})$ for Blood, as these resulted in the best accuracies.
Tab.~\ref{tab:datasets} and Fig.~\ref{fig:datasets_rec} display the results of these experiments.
Fig.~\ref{fig:takeaways} again visually illustrates the trade-off between model utility and privacy for all considered datasets.

The majority of results again confirm the previous findings for other datasets.
Our proposed CVB significantly decreases gradient leakage for all architectures.
Attack success for the unprotected CNN is larger than $96\%$ for all datasets.
If we add our proposed CVB to the model, the ASR drops to $0\%$ for most datasets.
The significant decrease in SSIM reflect this decrease in reconstruction quality accordingly.
Only for the Pneumonia dataset, we observed a ASR of $7.81\%$
We attribute this to the lower complexity of the dataset, \ie, gray-scale vs. colored images.
Model utility improved for most datasets, when our CVB was applied.
Only for the Derma dataset we observed a minor decrease in accuracy of $0.37\%$.

\section{Conclusion}
\label{sec:conclusion}
In this paper we analyzed gradients during gradient inversion attacks and revealed how PRECODE counters such attacks by design.
We demonstrated that the privacy-preserving properties of PRECODE arise from the stochastic gradients induced by the use of variational modeling.
Next, we formulated an attack that purposefully ignores the gradients of these layers during the attack to disable PRECODE's privacy preserving capabilities and show that training data can be successfully reconstructed.
PRECODE forces an attacker to omit previously usable gradient information during attack optimization.
Hence, to maintain privacy protection the stochastic module has to be placed early into the model.
For PRECODE, however, this results in reduced model utility and an exploding number of additional model parameters, \ie, higher computational and communication overhead.
To diminish these side effects we have proposed a novel Convolutional Variational Bottleneck (CVB) that can be placed early into the model without suffering from these side effects.
We show that CVB (1) improves privacy against iterative GI attacks, (2) requires fewer additional parameters than PRECODE and therefore less computational and communication resources during federated learning, and (3) can improve model utility.
Our findings are supported by systematic experiments on three seminal model architectures trained on six image classification datasets and attacked by three common GI attacks.

\bibliographystyle{IEEEtran}
\bibliography{bibliography.bib}

\clearpage

\appendix

\section*{Defending against analytical attacks}
\label{sec:apx_analytic_attack}

Since the input to any fully connected layer in a neural network can be analytically reconstructed~\cite{aono2017privacy, geiping2020inverting}, Multi Layer Perceptrons (MLP) are -- even with the protection of PRECODE -- vulnerable to gradient inversion~\cite{balunovic2021bayesian}. 
Given the gradients of the loss $\mathcal{L}$ with regards to the weights $\Theta^{(l)}_{W}$ and the bias $\Theta^{(l)}_{b}$ of a fully connected layer $l$, the input $x$ to that layer can be perfectly reconstructed by calculating:
\begin{equation}
    \label{eq:MLPatt}
    x^T = \left( \frac{d\mathcal{L}}{d\Theta^{(l)}_{b_i}} \right)^{-1} \cdot \frac{d\mathcal{L}}{d\Theta^{(l)}_{W_i}},
\end{equation}
where $i$ refers to the line of $\Theta^{(l)}_{W}$ and $\Theta^{(l)}_{b}$.
Further analysis of analytical reconstruction techniques is presented in~\cite{wang2020sapag, fan2020rethinking, zhu2020r}.
Note that $\frac{d\mathcal{L}}{d\Theta^{(l)}_{b}}$ requires to have at least one non-zero entry, which experimentally seems to always be the case~\cite{balunovic2021bayesian}.

However, we propose to remove the bias weights from all fully connected layers to entirely mitigate this problem.
We empirically analyzed the impact of this removal on the performance of models in a federated training scenario as described in the experimental setup.
As model architecture for this experiment we naturally consider a MLP which is composed of four fully connected layers with $1024$ neurons, batch normalization and ReLU activation.
The number of neurons in the last fully connected output layer with softmax activation is equal to the number of classes within a dataset.
The model is trained with and without PRECODE and both with enabled and disabled biases for the CIFAR-10 dataset.

Tab.~\ref{tab:bias} summarizes the test accuracies for a simple MLP with/without bias and with/without PRECODE. 
As removing the bias weights from the models only marginally affects model performance, it poses a simple yet effective defense against analytical attack.
Furthermore, we did not observe any effects on the success of iterative gradient inversion attacks when biases were disabled.

\begin{table}[!h]
\centering
\caption{Test accuracies for a MLP trained in a federated manner with and without bias weights and PRECODE.}
\label{tab:bias}
\begin{tabular}{l|cc}
\toprule
 Accuracy [$\%$] $\uparrow$& Bias & No Bias \\
\midrule
Baseline   & 59.91         & 59.98            \\
PRECODE    & 59.80         & 59.67            \\
\bottomrule
\end{tabular}
\end{table}


\section*{Code and Data}
\label{sec:code_apx}
Our experiments are based on the PyTorch implementation of IG\footnote[1]{\url{https://github.com/JonasGeiping/invertinggradients}}~\cite{geiping2020inverting}.
An implementation of our proposed targeted attack and the Convolutional Variational Bottleneck (CVB) is available on GitHub\footnote[2]{Published upon acceptance}.

All of our experiments were performed on publicly available datasets: MNIST~\cite{lecun1998gradient}, CIFAR-10~\cite{krizhevsky2009learning} and MedMNIST~\cite{yang2023medmnist}.
Victim client datasets for reconstruction are randomly sampled from the training data split of a single client.
We make these victim dataset available in our public code repository\footnotemark[2].

We performed our experiments on NVIDIA GeForce RTX 2080 Ti GPU, Intel(R) Xeon(R) Silver 4114 CPU and $128$ GB of working memory under \textit{Linux-5.4.0-91-generic-x86\_64-with-glibc2.31} OS.
We use Python version $3.11.3$ and PyTorch version $1.13.1$.
A more detailed list of the names and versions of the libraries and frameworks used is also available in our public code repository\footnotemark[2].

\section*{Additional Metrics and Experimental Results}
\label{sec:add_exp}
Besides \textit{Structural Similarity} (SSIM)~\cite{wang2004image} and \textit{Attack Success Rate} (ASR)~\cite{wagner2018technical}, we report \textit{Mean Squared Error} (MSE), \textit{Peak Signal-to-Noise Ratio} (PSNR), and a \textit{Learned Perceptual Image Patch Similarity} (LPIPS)~\cite{zhang2018unreasonable} to measure the reconstruction quality of the reconstructed images.
Lower SSIM, ASR and PSNR as well as higher MSE and LPIPS indicates lower reconstruction quality and therefore better privacy.
For SSIM, PSNR, MSE and LPIPS we report the average and standard deviation across the $128$ samples of each victim dataset.
ASR aggregates the ratio of successful attacks across all $128$ samples of each victim dataset.
An attack is considered successful if an SSIM of $0.5$ or greater is achieved.
Tables~\ref{tab:apx_ablation}-~\ref{tab:apx_datasets} shows more detailed results of the systematic empirical evaluation of our proposed CVB.

\begin{table*}
\centering
\caption{
Model utility and privacy metrics for a CNN without and with our proposed CVB defense mechanism trained CIFAR-10. 
When PRECODE or our CVB defense is used, our proposed targeted attack is applied.
Arrows indicate direction of improvement. 
Bold and italic formatting highlight best and worst results respectively.
}
\label{tab:apx_ablation}
\begin{tabular}{c|c|c|c|c|c|c|c|c|c|c}
\toprule
Defense & $P$ & $k_E$ & $s_E$ & $\beta$ & SSIM $\downarrow$ & ASR [\%] $\downarrow$ & PSNR $\downarrow$ & MSE $\uparrow$ & LPIPS $\uparrow$ & Accuracy [\%] $\uparrow$ \\
\midrule
None & - & - & - & - & {\cellcolor[HTML]{E0422F}} \color[HTML]{F1F1F1} \itshape 0.87 ($\pm$0.11) & {\cellcolor[HTML]{B30D26}} \color[HTML]{F1F1F1} \itshape 96.88 & {\cellcolor[HTML]{A50026}} \color[HTML]{F1F1F1} \itshape 21.09 ($\pm$2.88) & {\cellcolor[HTML]{D22B27}} \color[HTML]{F1F1F1} \itshape 0.01 ($\pm$0.01) & {\cellcolor[HTML]{FDC574}} \color[HTML]{000000} \itshape 0.20 ($\pm$0.09) & {\cellcolor[HTML]{CBE982}} \color[HTML]{000000} 62.53 ($\pm$0.16) \\

\midrule

\multirow[c]{7}{*}{Ours} & 1 & \multirow[c]{7}{*}{1} & \multirow[c]{7}{*}{$\frac{1}{2}$} & \multirow[c]{7}{*}{$10^{-1}$} & {\cellcolor[HTML]{30A356}} \color[HTML]{F1F1F1} 0.13 ($\pm$0.06) & {\cellcolor[HTML]{006837}} \color[HTML]{F1F1F1} \bfseries 0.00 & {\cellcolor[HTML]{F2FAAE}} \color[HTML]{000000} \bfseries 9.82 ($\pm$1.45) & {\cellcolor[HTML]{006837}} \color[HTML]{F1F1F1} \bfseries 0.11 ($\pm$0.04) & {\cellcolor[HTML]{0D8044}} \color[HTML]{F1F1F1} 0.55 ($\pm$0.06) & {\cellcolor[HTML]{D9EF8B}} \color[HTML]{000000} 62.01 ($\pm$1.66) \\
 & 2 &  &  &   & {\cellcolor[HTML]{7FC866}} \color[HTML]{000000} 0.24 ($\pm$0.09) & {\cellcolor[HTML]{006837}} \color[HTML]{F1F1F1} \bfseries 0.00 & {\cellcolor[HTML]{F5FBB2}} \color[HTML]{000000} 10.05 ($\pm$1.38) & {\cellcolor[HTML]{17934E}} \color[HTML]{F1F1F1} 0.10 ($\pm$0.03) & {\cellcolor[HTML]{006837}} \color[HTML]{F1F1F1} \bfseries 0.58 ($\pm$0.08) & {\cellcolor[HTML]{82C966}} \color[HTML]{000000} \bfseries 65.10 ($\pm$1.11) \\
 & 3 &   &  &   & {\cellcolor[HTML]{EFF8AA}} \color[HTML]{000000} 0.46 ($\pm$0.17) & {\cellcolor[HTML]{F7FCB4}} \color[HTML]{000000} 47.66 & {\cellcolor[HTML]{FED27F}} \color[HTML]{000000} 13.19 ($\pm$2.60) & {\cellcolor[HTML]{EEF8A8}} \color[HTML]{000000} 0.06 ($\pm$0.03) & {\cellcolor[HTML]{51B35E}} \color[HTML]{F1F1F1} 0.48 ($\pm$0.10) & {\cellcolor[HTML]{A7D96B}} \color[HTML]{000000} 63.93 ($\pm$0.59) \\
 & (1,2) &   &  &   & {\cellcolor[HTML]{30A356}} \color[HTML]{F1F1F1} 0.13 ($\pm$0.07) & {\cellcolor[HTML]{006837}} \color[HTML]{F1F1F1} \bfseries 0.00 & {\cellcolor[HTML]{FFF8B4}} \color[HTML]{000000} 10.96 ($\pm$1.73) & {\cellcolor[HTML]{57B65F}} \color[HTML]{F1F1F1} 0.09 ($\pm$0.04) & {\cellcolor[HTML]{279F53}} \color[HTML]{F1F1F1} 0.51 ($\pm$0.06) & {\cellcolor[HTML]{D9EF8B}} \color[HTML]{000000} 61.96 ($\pm$1.19) \\
 & (1,3) &   &  &   & {\cellcolor[HTML]{279F53}} \color[HTML]{F1F1F1} \bfseries 0.12 ($\pm$0.07) & {\cellcolor[HTML]{006837}} \color[HTML]{F1F1F1} \bfseries 0.00 & {\cellcolor[HTML]{FDFEBC}} \color[HTML]{000000} 10.45 ($\pm$1.79) & {\cellcolor[HTML]{17934E}} \color[HTML]{F1F1F1} 0.10 ($\pm$0.05) & {\cellcolor[HTML]{16914D}} \color[HTML]{F1F1F1} 0.53 ($\pm$0.05) & {\cellcolor[HTML]{EFF8AA}} \color[HTML]{000000} \itshape 60.79 ($\pm$0.53) \\
 & (2,3) &   &  &   & {\cellcolor[HTML]{78C565}} \color[HTML]{000000} 0.23 ($\pm$0.10) & {\cellcolor[HTML]{006837}} \color[HTML]{F1F1F1} \bfseries 0.00 & {\cellcolor[HTML]{FFFCBA}} \color[HTML]{000000} 10.75 ($\pm$1.43) & {\cellcolor[HTML]{57B65F}} \color[HTML]{F1F1F1} 0.09 ($\pm$0.03) & {\cellcolor[HTML]{04703B}} \color[HTML]{F1F1F1} 0.57 ($\pm$0.06) & {\cellcolor[HTML]{84CA66}} \color[HTML]{000000} 65.05 ($\pm$0.36) \\
 & (1,2,3) &   &  &   & {\cellcolor[HTML]{36A657}} \color[HTML]{F1F1F1} 0.14 ($\pm$0.09) & {\cellcolor[HTML]{006837}} \color[HTML]{F1F1F1} \bfseries 0.00 & {\cellcolor[HTML]{FEEDA1}} \color[HTML]{000000} 11.78 ($\pm$2.08) & {\cellcolor[HTML]{C7E77F}} \color[HTML]{000000} 0.07 ($\pm$0.04) & {\cellcolor[HTML]{279F53}} \color[HTML]{F1F1F1} 0.51 ($\pm$0.06) & {\cellcolor[HTML]{EFF8AA}} \color[HTML]{000000} 60.80 ($\pm$0.66) \\

\midrule 

\multirow[c]{4}{*}{Ours} & \multirow[c]{4}{*}{1} & 1 & \multirow[c]{4}{*}{$\frac{1}{2}$} & \multirow[c]{4}{*}{$10^{-1}$} & {\cellcolor[HTML]{30A356}} \color[HTML]{F1F1F1} \bfseries 0.13 ($\pm$0.06) & {\cellcolor[HTML]{006837}} \color[HTML]{F1F1F1} \bfseries 0.00 & {\cellcolor[HTML]{F2FAAE}} \color[HTML]{000000} 9.82 ($\pm$1.45) & {\cellcolor[HTML]{15904C}} \color[HTML]{F1F1F1} 0.11 ($\pm$0.04) & {\cellcolor[HTML]{0D8044}} \color[HTML]{F1F1F1} 0.55 ($\pm$0.06) & {\cellcolor[HTML]{D9EF8B}} \color[HTML]{000000} \itshape 62.01 ($\pm$1.66) \\
 &  & 3 &  &   & {\cellcolor[HTML]{5DB961}} \color[HTML]{F1F1F1} 0.19 ($\pm$0.07) & {\cellcolor[HTML]{006837}} \color[HTML]{F1F1F1} \bfseries 0.00 & {\cellcolor[HTML]{F5FBB2}} \color[HTML]{000000} 9.97 ($\pm$1.48) & {\cellcolor[HTML]{15904C}} \color[HTML]{F1F1F1} 0.11 ($\pm$0.04) & {\cellcolor[HTML]{0D8044}} \color[HTML]{F1F1F1} 0.55 ($\pm$0.06) & {\cellcolor[HTML]{69BE63}} \color[HTML]{F1F1F1} 65.88 ($\pm$1.11) \\
 &  & 5 &  &   & {\cellcolor[HTML]{6BBF64}} \color[HTML]{000000} 0.21 ($\pm$0.07) & {\cellcolor[HTML]{006837}} \color[HTML]{F1F1F1} \bfseries 0.00 & {\cellcolor[HTML]{ECF7A6}} \color[HTML]{000000} \bfseries 9.51 ($\pm$1.49) & {\cellcolor[HTML]{006837}} \color[HTML]{F1F1F1} \bfseries 0.12 ($\pm$0.04) & {\cellcolor[HTML]{006837}} \color[HTML]{F1F1F1} \bfseries 0.58 ($\pm$0.06) & {\cellcolor[HTML]{39A758}} \color[HTML]{F1F1F1} 67.17 ($\pm$0.58) \\
 &  & 7 &  &   & {\cellcolor[HTML]{78C565}} \color[HTML]{000000} 0.23 ($\pm$0.08) & {\cellcolor[HTML]{006837}} \color[HTML]{F1F1F1} \bfseries 0.00 & {\cellcolor[HTML]{F5FBB2}} \color[HTML]{000000} 10.03 ($\pm$1.56) & {\cellcolor[HTML]{15904C}} \color[HTML]{F1F1F1} 0.11 ($\pm$0.04) & {\cellcolor[HTML]{04703B}} \color[HTML]{F1F1F1} 0.57 ($\pm$0.06) & {\cellcolor[HTML]{16914D}} \color[HTML]{F1F1F1} \bfseries 68.23 ($\pm$0.19) \\

\midrule 

\multirow[c]{6}{*}{Ours} & \multirow[c]{6}{*}{1} & \multirow[c]{6}{*}{5} & $\frac{1}{16}$ & \multirow[c]{6}{*}{$10^{-1}$} & {\cellcolor[HTML]{199750}} \color[HTML]{F1F1F1} \bfseries 0.10 ($\pm$0.06) & {\cellcolor[HTML]{006837}} \color[HTML]{F1F1F1} \bfseries 0.00 & {\cellcolor[HTML]{DDF191}} \color[HTML]{000000} 8.70 ($\pm$1.51) & {\cellcolor[HTML]{006837}} \color[HTML]{F1F1F1} \bfseries 0.14 ($\pm$0.05) & {\cellcolor[HTML]{04703B}} \color[HTML]{F1F1F1} 0.61 ($\pm$0.06) & {\cellcolor[HTML]{FBA35C}} \color[HTML]{000000} \itshape 55.63 ($\pm$0.39) \\
 &  &  & $\frac{1}{4}$ &   & {\cellcolor[HTML]{3FAA59}} \color[HTML]{F1F1F1} 0.15 ($\pm$0.06) & {\cellcolor[HTML]{006837}} \color[HTML]{F1F1F1} \bfseries 0.00 & {\cellcolor[HTML]{DCF08F}} \color[HTML]{000000} \bfseries 8.59 ($\pm$1.17) & {\cellcolor[HTML]{006837}} \color[HTML]{F1F1F1} \bfseries 0.14 ($\pm$0.04) & {\cellcolor[HTML]{006837}} \color[HTML]{F1F1F1} \bfseries 0.62 ($\pm$0.06) & {\cellcolor[HTML]{7DC765}} \color[HTML]{000000} 65.24 ($\pm$0.31) \\
 & &  & $\frac{1}{2}$ &   & {\cellcolor[HTML]{6BBF64}} \color[HTML]{000000} 0.21 ($\pm$0.07) & {\cellcolor[HTML]{006837}} \color[HTML]{F1F1F1} \bfseries 0.00 & {\cellcolor[HTML]{ECF7A6}} \color[HTML]{000000} 9.51 ($\pm$1.49) & {\cellcolor[HTML]{39A758}} \color[HTML]{F1F1F1} 0.12 ($\pm$0.04) & {\cellcolor[HTML]{108647}} \color[HTML]{F1F1F1} 0.58 ($\pm$0.06) & {\cellcolor[HTML]{39A758}} \color[HTML]{F1F1F1} 67.17 ($\pm$0.58) \\
 &  &  & $1$ &   & {\cellcolor[HTML]{A5D86A}} \color[HTML]{000000} 0.30 ($\pm$0.08) & {\cellcolor[HTML]{036E3A}} \color[HTML]{F1F1F1} 1.56 & {\cellcolor[HTML]{FFF0A6}} \color[HTML]{000000} 11.60 ($\pm$1.83) & {\cellcolor[HTML]{E3F399}} \color[HTML]{000000} 0.08 ($\pm$0.03) & {\cellcolor[HTML]{48AE5C}} \color[HTML]{F1F1F1} 0.52 ($\pm$0.06) & {\cellcolor[HTML]{18954F}} \color[HTML]{F1F1F1} 68.05 ($\pm$0.63) \\
 &  &  & $2$ &   & {\cellcolor[HTML]{CFEB85}} \color[HTML]{000000} 0.38 ($\pm$0.11) & {\cellcolor[HTML]{2DA155}} \color[HTML]{F1F1F1} 12.50 & {\cellcolor[HTML]{FECC7B}} \color[HTML]{000000} 13.45 ($\pm$1.80) & {\cellcolor[HTML]{FECA79}} \color[HTML]{000000} 0.05 ($\pm$0.03) & {\cellcolor[HTML]{8CCD67}} \color[HTML]{000000} 0.46 ($\pm$0.07) & {\cellcolor[HTML]{0F8446}} \color[HTML]{F1F1F1} 68.79 ($\pm$1.32) \\
 &  &  & $4$ &   & {\cellcolor[HTML]{E8F59F}} \color[HTML]{000000} 0.44 ($\pm$0.10) & {\cellcolor[HTML]{8ECF67}} \color[HTML]{000000} 26.56 & {\cellcolor[HTML]{FDB163}} \color[HTML]{000000} 14.59 ($\pm$1.76) & {\cellcolor[HTML]{FCA55D}} \color[HTML]{000000} 0.04 ($\pm$0.02) & {\cellcolor[HTML]{B1DE71}} \color[HTML]{000000} 0.42 ($\pm$0.07) & {\cellcolor[HTML]{0D8044}} \color[HTML]{F1F1F1} \bfseries 68.93 ($\pm$0.58) \\

\midrule 

\multirow[c]{5}{*}{Ours} & \multirow[c]{5}{*}{1} & \multirow[c]{5}{*}{5} & \multirow[c]{5}{*}{$\frac{1}{2}$} & $10^{-1}$ & {\cellcolor[HTML]{6BBF64}} \color[HTML]{000000} 0.21 ($\pm$0.07) & {\cellcolor[HTML]{006837}} \color[HTML]{F1F1F1} \bfseries 0.00 & {\cellcolor[HTML]{ECF7A6}} \color[HTML]{000000} \bfseries 9.51 ($\pm$1.49) & {\cellcolor[HTML]{006837}} \color[HTML]{F1F1F1} \bfseries 0.12 ($\pm$0.04) & {\cellcolor[HTML]{04703B}} \color[HTML]{F1F1F1} 0.58 ($\pm$0.06) & {\cellcolor[HTML]{39A758}} \color[HTML]{F1F1F1} \bfseries 67.17 ($\pm$0.58) \\
 &  &  &  & $10^{-2}$ & {\cellcolor[HTML]{5DB961}} \color[HTML]{F1F1F1} \bfseries 0.19 ($\pm$0.07) & {\cellcolor[HTML]{006837}} \color[HTML]{F1F1F1} \bfseries 0.00 & {\cellcolor[HTML]{F1F9AC}} \color[HTML]{000000} 9.75 ($\pm$1.36) & {\cellcolor[HTML]{15904C}} \color[HTML]{F1F1F1} 0.11 ($\pm$0.04) & {\cellcolor[HTML]{006837}} \color[HTML]{F1F1F1} \bfseries 0.59 ($\pm$0.06) & {\cellcolor[HTML]{87CB67}} \color[HTML]{000000} 64.97 ($\pm$0.19) \\
 &  &  &  & $10^{-3}$ & {\cellcolor[HTML]{5DB961}} \color[HTML]{F1F1F1} \bfseries 0.19 ($\pm$0.07) & {\cellcolor[HTML]{006837}} \color[HTML]{F1F1F1} \bfseries 0.00 & {\cellcolor[HTML]{FBFDBA}} \color[HTML]{000000} 10.34 ($\pm$1.12) & {\cellcolor[HTML]{4BB05C}} \color[HTML]{F1F1F1} 0.10 ($\pm$0.02) & {\cellcolor[HTML]{04703B}} \color[HTML]{F1F1F1} 0.58 ($\pm$0.06) & {\cellcolor[HTML]{B1DE71}} \color[HTML]{000000} 63.55 ($\pm$0.50) \\
 &  &  &  & $10^{-4}$ & {\cellcolor[HTML]{66BD63}} \color[HTML]{F1F1F1} 0.20 ($\pm$0.07) & {\cellcolor[HTML]{006837}} \color[HTML]{F1F1F1} \bfseries 0.00 & {\cellcolor[HTML]{FBFDBA}} \color[HTML]{000000} 10.32 ($\pm$1.25) & {\cellcolor[HTML]{4BB05C}} \color[HTML]{F1F1F1} 0.10 ($\pm$0.03) & {\cellcolor[HTML]{04703B}} \color[HTML]{F1F1F1} 0.58 ($\pm$0.06) & {\cellcolor[HTML]{B7E075}} \color[HTML]{000000} 63.31 ($\pm$0.54) \\
 &  &  &  & $10^{-5}$ & {\cellcolor[HTML]{6BBF64}} \color[HTML]{000000} 0.21 ($\pm$0.07) & {\cellcolor[HTML]{016A38}} \color[HTML]{F1F1F1} 0.78 & {\cellcolor[HTML]{FDFEBC}} \color[HTML]{000000} 10.43 ($\pm$1.24) & {\cellcolor[HTML]{84CA66}} \color[HTML]{000000} 0.09 ($\pm$0.03) & {\cellcolor[HTML]{08773F}} \color[HTML]{F1F1F1} 0.57 ($\pm$0.07) & {\cellcolor[HTML]{B1DE71}} \color[HTML]{000000} 63.56 ($\pm$0.66) \\

\bottomrule
\end{tabular}
\end{table*}

\begin{table*}
\centering
\caption{
Privacy metrics for a CNN without and with our proposed CVB defense mechanism on MNIST and CIFAR-10. 
The gradients are attacked with different GI attacks. 
When our defense is used, our proposed targeted attack is applied.
Arrows indicate direction of improvement. 
Bold and italic formatting highlight best and worst results respectively.
}
\label{tab:apx_attacks}
\begin{tabular}{c|c|c|c|c|c|c|c|c|c|c|c}
\toprule
 & Defense & Attack & $P$ & $k_E$ & $s_E$ & $\beta$ & SSIM $\downarrow$ & ASR [\%] $\downarrow$ & PSNR $\downarrow$ & MSE $\uparrow$ & LPIPS $\uparrow$ \\
\midrule
\multirow[c]{6}{*}{\rotatebox{90}{MNIST}} & \multirow[c]{3}{*}{None} & iDLG & \multirow[c]{3}{*}{-} & \multirow[c]{3}{*}{-} & \multirow[c]{3}{*}{-} & \multirow[c]{3}{*}{-} & {\cellcolor[HTML]{F8864F}} \color[HTML]{F1F1F1} 0.76 ($\pm$0.18) & {\cellcolor[HTML]{CE2827}} \color[HTML]{F1F1F1} 91.41 & {\cellcolor[HTML]{F8FCB6}} \color[HTML]{000000} 15.98 ($\pm$2.64) & {\cellcolor[HTML]{DC3B2C}} \color[HTML]{F1F1F1} 0.03 ($\pm$0.04) & {\cellcolor[HTML]{FFF6B0}} \color[HTML]{000000} 0.40 ($\pm$0.13) \\
 &  & CPL & & & & & {\cellcolor[HTML]{AFDD70}} \color[HTML]{000000} 0.32 ($\pm$0.17) & {\cellcolor[HTML]{2AA054}} \color[HTML]{F1F1F1} 12.40 & {\cellcolor[HTML]{78C565}} \color[HTML]{000000} 7.63 ($\pm$1.18) & {\cellcolor[HTML]{98D368}} \color[HTML]{000000} 0.18 ($\pm$0.05) & {\cellcolor[HTML]{C1E57B}} \color[HTML]{000000} 0.55 ($\pm$0.17) \\
 &  & IG & & & & & {\cellcolor[HTML]{BD1726}} \color[HTML]{F1F1F1} \itshape 0.95 ($\pm$0.06) & {\cellcolor[HTML]{A50026}} \color[HTML]{F1F1F1} \itshape 100.00 & {\cellcolor[HTML]{A50026}} \color[HTML]{F1F1F1} \itshape 33.13 ($\pm$7.63) & {\cellcolor[HTML]{A50026}} \color[HTML]{F1F1F1} \itshape 0.00 ($\pm$0.00) & {\cellcolor[HTML]{B71126}} \color[HTML]{F1F1F1} \itshape 0.03 ($\pm$0.06) \\
 \cmidrule{2-12}
 & \multirow[c]{3}{*}{Ours} & iDLG & \multirow[c]{3}{*}{1} & \multirow[c]{3}{*}{5} & \multirow[c]{3}{*}{$\frac{1}{2}$} & \multirow[c]{3}{*}{$10^{-1}$} & {\cellcolor[HTML]{006837}} \color[HTML]{F1F1F1} \bfseries 0.00 ($\pm$0.02) & {\cellcolor[HTML]{006837}} \color[HTML]{F1F1F1} \bfseries 0.00 & {\cellcolor[HTML]{8CCD67}} \color[HTML]{000000} 8.67 ($\pm$0.69) & {\cellcolor[HTML]{E8F59F}} \color[HTML]{000000} 0.14 ($\pm$0.02) & {\cellcolor[HTML]{097940}} \color[HTML]{F1F1F1} 0.82 ($\pm$0.03) \\
 &  & CPL & & &  & & {\cellcolor[HTML]{006837}} \color[HTML]{F1F1F1} \bfseries 0.00 ($\pm$0.02) & {\cellcolor[HTML]{006837}} \color[HTML]{F1F1F1} \bfseries 0.00 & {\cellcolor[HTML]{5AB760}} \color[HTML]{F1F1F1} \bfseries 6.13 ($\pm$0.54) & {\cellcolor[HTML]{006837}} \color[HTML]{F1F1F1} \bfseries 0.25 ($\pm$0.03) & {\cellcolor[HTML]{006837}} \color[HTML]{F1F1F1} \bfseries 0.85 ($\pm$0.03) \\
 &  & IG & & &  & & {\cellcolor[HTML]{A0D669}} \color[HTML]{000000} 0.29 ($\pm$0.09) & {\cellcolor[HTML]{036E3A}} \color[HTML]{F1F1F1} 1.56 & {\cellcolor[HTML]{73C264}} \color[HTML]{000000} 7.35 ($\pm$1.60) & {\cellcolor[HTML]{66BD63}} \color[HTML]{F1F1F1} 0.20 ($\pm$0.07) & {\cellcolor[HTML]{84CA66}} \color[HTML]{000000} 0.64 ($\pm$0.05) \\

\midrule

\multirow[c]{6}{*}{\rotatebox{90}{CIFAR-10}} & \multirow[c]{3}{*}{None} & iDLG & \multirow[c]{3}{*}{-} & \multirow[c]{3}{*}{-} & \multirow[c]{3}{*}{-} & \multirow[c]{3}{*}{-} & {\cellcolor[HTML]{E5F49B}} \color[HTML]{000000} 0.43 ($\pm$0.23) & {\cellcolor[HTML]{F7FCB4}} \color[HTML]{000000} 47.66 & {\cellcolor[HTML]{FEEDA1}} \color[HTML]{000000} 11.76 ($\pm$2.64) & {\cellcolor[HTML]{B7E075}} \color[HTML]{000000} 0.08 ($\pm$0.06) & {\cellcolor[HTML]{60BA62}} \color[HTML]{F1F1F1} 0.54 ($\pm$0.10) \\
 &  & CPL & & & & & {\cellcolor[HTML]{AFDD70}} \color[HTML]{000000} 0.32 ($\pm$0.20) & {\cellcolor[HTML]{78C565}} \color[HTML]{000000} 22.66 & {\cellcolor[HTML]{FEE695}} \color[HTML]{000000} 12.25 ($\pm$1.84) & {\cellcolor[HTML]{DFF293}} \color[HTML]{000000} 0.07 ($\pm$0.03) & {\cellcolor[HTML]{54B45F}} \color[HTML]{F1F1F1} 0.55 ($\pm$0.09) \\
 &  & IG & & & & & {\cellcolor[HTML]{E0422F}} \color[HTML]{F1F1F1} \itshape 0.87 ($\pm$0.11) & {\cellcolor[HTML]{B30D26}} \color[HTML]{F1F1F1} \itshape 96.88 & {\cellcolor[HTML]{A50026}} \color[HTML]{F1F1F1} \itshape 21.09 ($\pm$2.88) & {\cellcolor[HTML]{CE2827}} \color[HTML]{F1F1F1} \itshape 0.01 ($\pm$0.01) & {\cellcolor[HTML]{FDAD60}} \color[HTML]{000000} \itshape 0.20 ($\pm$0.09) \\
 \cmidrule{2-12}
 & \multirow[c]{3}{*}{Ours} & iDLG & \multirow[c]{3}{*}{1} & \multirow[c]{3}{*}{5} & \multirow[c]{3}{*}{$\frac{1}{2}$} & \multirow[c]{3}{*}{$10^{-1}$} & {\cellcolor[HTML]{026C39}} \color[HTML]{F1F1F1} \bfseries 0.01 ($\pm$0.01) & {\cellcolor[HTML]{006837}} \color[HTML]{F1F1F1} \bfseries 0.00 & {\cellcolor[HTML]{E5F49B}} \color[HTML]{000000} \bfseries 9.14 ($\pm$0.91) & {\cellcolor[HTML]{006837}} \color[HTML]{F1F1F1} \bfseries 0.12 ($\pm$0.03) & {\cellcolor[HTML]{006837}} \color[HTML]{F1F1F1} \bfseries 0.67 ($\pm$0.04) \\
 &  & CPL & & &  & & {\cellcolor[HTML]{07753E}} \color[HTML]{F1F1F1} 0.03 ($\pm$0.02) & {\cellcolor[HTML]{006837}} \color[HTML]{F1F1F1} \bfseries 0.00 & {\cellcolor[HTML]{FFFEBE}} \color[HTML]{000000} 10.55 ($\pm$1.32) & {\cellcolor[HTML]{84CA66}} \color[HTML]{000000} 0.09 ($\pm$0.03) & {\cellcolor[HTML]{0B7D42}} \color[HTML]{F1F1F1} 0.64 ($\pm$0.04) \\
 &  & IG & & &  & & {\cellcolor[HTML]{6BBF64}} \color[HTML]{000000} 0.21 ($\pm$0.08) & {\cellcolor[HTML]{006837}} \color[HTML]{F1F1F1} \bfseries 0.00 & {\cellcolor[HTML]{ECF7A6}} \color[HTML]{000000} 9.51 ($\pm$1.49) & {\cellcolor[HTML]{006837}} \color[HTML]{F1F1F1} \bfseries 0.12 ($\pm$0.04) & {\cellcolor[HTML]{33A456}} \color[HTML]{F1F1F1} 0.58 ($\pm$0.06) \\
\bottomrule
\end{tabular}
\end{table*}

\begin{table*}
\centering
\caption{
Model utility and privacy metrics for a CNN without and with our proposed CVB defense mechanism trained on MNIST and CIFAR-10. 
The gradients are attacked for different defense mechanisms.
Parameters indicate clipping threshold and noise multiplier $(C, \sigma)$ for DP; pruning ratio $p$ for GC; position and bottleneck size $(P,K)$ for PRECODE, position, kernel size, bottleneck scale and loss weight $(P, k_E, s_E, \beta)$ for our proposed CVB.
When PRECODE or our CVB defense is used, our proposed targeted attack is applied (cf. Section~\ref{sec:anti_precode}).
Arrows indicate direction of improvement. 
Bold and italic formatting highlight best and worst results respectively.}
\label{tab:apx_defenses}
\begin{tabular}{c|c|c|c|c|c|c|c|c}
\toprule
 & Defense & Parameters & SSIM $\downarrow$ & ASR [\%] $\downarrow$ & PSNR $\downarrow$ & MSE $\uparrow$ & LPIPS $\uparrow$ & Accuracy [\%] $\uparrow$ \\
\midrule
\multirow[c]{17}{*}{\rotatebox{90}{MNIST}} & None & - & {\cellcolor[HTML]{BD1726}} \color[HTML]{F1F1F1} \itshape 0.95 ($\pm$0.06) & {\cellcolor[HTML]{A50026}} \color[HTML]{F1F1F1} \itshape 100.00 & {\cellcolor[HTML]{A50026}} \color[HTML]{F1F1F1} \itshape 33.13 ($\pm$7.63) & {\cellcolor[HTML]{A50026}} \color[HTML]{F1F1F1} \itshape 0.00 ($\pm$0.00) & {\cellcolor[HTML]{B91326}} \color[HTML]{F1F1F1} \itshape 0.03 ($\pm$0.06) & {\cellcolor[HTML]{006837}} \color[HTML]{F1F1F1} 99.10 ($\pm$0.04) \\
 \cmidrule{2-9}
 
 & \multirow[c]{4}{*}{DP} & $(20, 1)$ & {\cellcolor[HTML]{006837}} \color[HTML]{F1F1F1} \bfseries 0.00 ($\pm$0.03) & {\cellcolor[HTML]{006837}} \color[HTML]{F1F1F1} \bfseries 0.00 & {\cellcolor[HTML]{75C465}} \color[HTML]{000000} 7.43 ($\pm$0.61) & {\cellcolor[HTML]{57B65F}} \color[HTML]{F1F1F1} 0.18 ($\pm$0.03) & {\cellcolor[HTML]{006837}} \color[HTML]{F1F1F1} \bfseries 0.74 ($\pm$0.04) & {\cellcolor[HTML]{006837}} \color[HTML]{F1F1F1} 94.73 ($\pm$0.12) \\
 &  & $(20, 10^{-1})$ & {\cellcolor[HTML]{026C39}} \color[HTML]{F1F1F1} 0.01 ($\pm$0.03) & {\cellcolor[HTML]{006837}} \color[HTML]{F1F1F1} \bfseries 0.00 & {\cellcolor[HTML]{78C565}} \color[HTML]{000000} 7.58 ($\pm$0.63) & {\cellcolor[HTML]{57B65F}} \color[HTML]{F1F1F1} 0.18 ($\pm$0.03) & {\cellcolor[HTML]{006837}} \color[HTML]{F1F1F1} \bfseries 0.74 ($\pm$0.04) & {\cellcolor[HTML]{006837}} \color[HTML]{F1F1F1} 98.17 ($\pm$0.07) \\
 &  & $(20, 10^{-2})$ & {\cellcolor[HTML]{148E4B}} \color[HTML]{F1F1F1} 0.08 ($\pm$0.04) & {\cellcolor[HTML]{006837}} \color[HTML]{F1F1F1} \bfseries 0.00 & {\cellcolor[HTML]{89CC67}} \color[HTML]{000000} 8.52 ($\pm$0.63) & {\cellcolor[HTML]{C7E77F}} \color[HTML]{000000} 0.14 ($\pm$0.02) & {\cellcolor[HTML]{006837}} \color[HTML]{F1F1F1} \bfseries 0.74 ($\pm$0.04) & {\cellcolor[HTML]{006837}} \color[HTML]{F1F1F1} 98.87 ($\pm$0.08) \\
 &  & $(20, 10^{-3})$ & {\cellcolor[HTML]{FECC7B}} \color[HTML]{000000} 0.64 ($\pm$0.11) & {\cellcolor[HTML]{E44C34}} \color[HTML]{F1F1F1} 85.16 & {\cellcolor[HTML]{FFF3AC}} \color[HTML]{000000} 17.83 ($\pm$1.46) & {\cellcolor[HTML]{D22B27}} \color[HTML]{F1F1F1} 0.02 ($\pm$0.01) & {\cellcolor[HTML]{E0F295}} \color[HTML]{000000} 0.43 ($\pm$0.09) & {\cellcolor[HTML]{006837}} \color[HTML]{F1F1F1} 98.97 ($\pm$0.06) \\
 \cmidrule{2-9}
 
& \multirow[c]{2}{*}{GC} & $0.9$ & {\cellcolor[HTML]{FED07E}} \color[HTML]{000000} 0.63 ($\pm$0.09) & {\cellcolor[HTML]{D83128}} \color[HTML]{F1F1F1} 89.84 & {\cellcolor[HTML]{FFF0A6}} \color[HTML]{000000} 18.12 ($\pm$1.91) & {\cellcolor[HTML]{D22B27}} \color[HTML]{F1F1F1} 0.02 ($\pm$0.01) & {\cellcolor[HTML]{E5F49B}} \color[HTML]{000000} 0.42 ($\pm$0.09) & {\cellcolor[HTML]{006837}} \color[HTML]{F1F1F1} 98.31 ($\pm$0.07) \\
 &  & $0.99$ & {\cellcolor[HTML]{7FC866}} \color[HTML]{000000} 0.24 ($\pm$0.05) & {\cellcolor[HTML]{006837}} \color[HTML]{F1F1F1} \bfseries 0.00 & {\cellcolor[HTML]{BFE47A}} \color[HTML]{000000} 11.57 ($\pm$1.14) & {\cellcolor[HTML]{FDB768}} \color[HTML]{000000} 0.07 ($\pm$0.02) & {\cellcolor[HTML]{06733D}} \color[HTML]{F1F1F1} 0.72 ($\pm$0.04) & {\cellcolor[HTML]{A50026}} \color[HTML]{F1F1F1} \itshape 23.08 ($\pm$9.39) \\
 \cmidrule{2-9}
 
 & \multirow[c]{7}{*}{PRECODE} & $(1, 8)$ & {\cellcolor[HTML]{ECF7A6}} \color[HTML]{000000} 0.45 ($\pm$0.10) & {\cellcolor[HTML]{BDE379}} \color[HTML]{000000} 34.38 & {\cellcolor[HTML]{ABDB6D}} \color[HTML]{000000} 10.26 ($\pm$1.94) & {\cellcolor[HTML]{FFF1A8}} \color[HTML]{000000} 0.10 ($\pm$0.05) & {\cellcolor[HTML]{118848}} \color[HTML]{F1F1F1} 0.69 ($\pm$0.07) & {\cellcolor[HTML]{006837}} \color[HTML]{F1F1F1} 98.30 ($\pm$0.08) \\
 &  & $(1, 16)$ & {\cellcolor[HTML]{ECF7A6}} \color[HTML]{000000} 0.45 ($\pm$0.09) & {\cellcolor[HTML]{A5D86A}} \color[HTML]{000000} 29.69 & {\cellcolor[HTML]{A9DA6C}} \color[HTML]{000000} 10.15 ($\pm$1.27) & {\cellcolor[HTML]{FFF1A8}} \color[HTML]{000000} 0.10 ($\pm$0.03) & {\cellcolor[HTML]{118848}} \color[HTML]{F1F1F1} 0.69 ($\pm$0.06) & {\cellcolor[HTML]{006837}} \color[HTML]{F1F1F1} 98.45 ($\pm$0.03) \\
 &  & $(1, 32)$ & {\cellcolor[HTML]{EFF8AA}} \color[HTML]{000000} 0.46 ($\pm$0.09) & {\cellcolor[HTML]{B7E075}} \color[HTML]{000000} 33.59 & {\cellcolor[HTML]{A9DA6C}} \color[HTML]{000000} 10.22 ($\pm$1.06) & {\cellcolor[HTML]{FFF1A8}} \color[HTML]{000000} 0.10 ($\pm$0.02) & {\cellcolor[HTML]{118848}} \color[HTML]{F1F1F1} 0.69 ($\pm$0.05) & {\cellcolor[HTML]{006837}} \color[HTML]{F1F1F1} 98.55 ($\pm$0.09) \\
 &  & $(1, 64)$ & {\cellcolor[HTML]{FFF8B4}} \color[HTML]{000000} 0.52 ($\pm$0.10) & {\cellcolor[HTML]{FEE18D}} \color[HTML]{000000} 59.38 & {\cellcolor[HTML]{C3E67D}} \color[HTML]{000000} 11.89 ($\pm$0.91) & {\cellcolor[HTML]{FDB768}} \color[HTML]{000000} 0.07 ($\pm$0.01) & {\cellcolor[HTML]{3FAA59}} \color[HTML]{F1F1F1} 0.63 ($\pm$0.05) & {\cellcolor[HTML]{006837}} \color[HTML]{F1F1F1} 98.42 ($\pm$0.06) \\
 &  & $(1, 128)$ & {\cellcolor[HTML]{FEE28F}} \color[HTML]{000000} 0.59 ($\pm$0.10) & {\cellcolor[HTML]{EC5C3B}} \color[HTML]{F1F1F1} 82.81 & {\cellcolor[HTML]{E6F59D}} \color[HTML]{000000} 14.47 ($\pm$1.11) & {\cellcolor[HTML]{EE613E}} \color[HTML]{F1F1F1} 0.04 ($\pm$0.01) & {\cellcolor[HTML]{89CC67}} \color[HTML]{000000} 0.55 ($\pm$0.06) & {\cellcolor[HTML]{006837}} \color[HTML]{F1F1F1} 98.49 ($\pm$0.10) \\
 &  & $(1, 256)$ & {\cellcolor[HTML]{FDB768}} \color[HTML]{000000} 0.68 ($\pm$0.10) & {\cellcolor[HTML]{D42D27}} \color[HTML]{F1F1F1} 90.62 & {\cellcolor[HTML]{FEEB9D}} \color[HTML]{000000} 18.67 ($\pm$1.25) & {\cellcolor[HTML]{BB1526}} \color[HTML]{F1F1F1} 0.01 ($\pm$0.00) & {\cellcolor[HTML]{EFF8AA}} \color[HTML]{000000} 0.40 ($\pm$0.07) & {\cellcolor[HTML]{006837}} \color[HTML]{F1F1F1} 98.37 ($\pm$0.18) \\
 &  & $(1, 512)$ & {\cellcolor[HTML]{F67A49}} \color[HTML]{F1F1F1} 0.78 ($\pm$0.09) & {\cellcolor[HTML]{A50026}} \color[HTML]{F1F1F1} \itshape 100.00 & {\cellcolor[HTML]{FA9656}} \color[HTML]{000000} 24.37 ($\pm$1.97) & {\cellcolor[HTML]{A50026}} \color[HTML]{F1F1F1} \itshape 0.00 ($\pm$0.00) & {\cellcolor[HTML]{FDBB6C}} \color[HTML]{000000} 0.24 ($\pm$0.06) & {\cellcolor[HTML]{006837}} \color[HTML]{F1F1F1} 98.22 ($\pm$0.07) \\
\cmidrule{2-9}

 & \textbf{Ours} & $(1, 5,\frac{1}{2}, 10^{-1})$ & {\cellcolor[HTML]{A0D669}} \color[HTML]{000000} 0.29 ($\pm$0.09) & {\cellcolor[HTML]{016A38}} \color[HTML]{F1F1F1} 0.78 & {\cellcolor[HTML]{73C264}} \color[HTML]{000000} 7.34 ($\pm$1.55) & {\cellcolor[HTML]{17934E}} \color[HTML]{F1F1F1} 0.20 ($\pm$0.07) & {\cellcolor[HTML]{33A456}} \color[HTML]{F1F1F1} 0.64 ($\pm$0.05) & {\cellcolor[HTML]{006837}} \color[HTML]{F1F1F1} \bfseries 99.15 ($\pm$0.05) \\
 
\midrule

\multirow[c]{17}{*}{\rotatebox{90}{CIFAR-10}} & None & - & {\cellcolor[HTML]{E0422F}} \color[HTML]{F1F1F1} \itshape 0.87 ($\pm$0.11) & {\cellcolor[HTML]{B30D26}} \color[HTML]{F1F1F1} 96.88 & {\cellcolor[HTML]{A50026}} \color[HTML]{F1F1F1} \itshape 21.09 ($\pm$2.88) & {\cellcolor[HTML]{C41E27}} \color[HTML]{F1F1F1} \itshape 0.01 ($\pm$0.01) & {\cellcolor[HTML]{FCAA5F}} \color[HTML]{000000} \itshape 0.20 ($\pm$0.09) & {\cellcolor[HTML]{CBE982}} \color[HTML]{000000} 62.53 ($\pm$0.16) \\
\cmidrule{2-9}
 & \multirow[c]{4}{*}{DP} & $(20, 1)$ & {\cellcolor[HTML]{026C39}} \color[HTML]{F1F1F1} \bfseries 0.01 ($\pm$0.02) & {\cellcolor[HTML]{006837}} \color[HTML]{F1F1F1} \bfseries 0.00 & {\cellcolor[HTML]{D1EC86}} \color[HTML]{000000} \bfseries 8.13 ($\pm$0.78) & {\cellcolor[HTML]{006837}} \color[HTML]{F1F1F1} \bfseries 0.16 ($\pm$0.03) & {\cellcolor[HTML]{006837}} \color[HTML]{F1F1F1} \bfseries 0.68 ($\pm$0.05) & {\cellcolor[HTML]{A50026}} \color[HTML]{F1F1F1} 49.87 ($\pm$1.19) \\
 &  & $(20, 10^{-1})$ & {\cellcolor[HTML]{026C39}} \color[HTML]{F1F1F1} \bfseries 0.01 ($\pm$0.02) & {\cellcolor[HTML]{006837}} \color[HTML]{F1F1F1} \bfseries 0.00 & {\cellcolor[HTML]{D3EC87}} \color[HTML]{000000} 8.16 ($\pm$0.77) & {\cellcolor[HTML]{006837}} \color[HTML]{F1F1F1} \bfseries 0.16 ($\pm$0.03) & {\cellcolor[HTML]{006837}} \color[HTML]{F1F1F1} \bfseries 0.68 ($\pm$0.04) & {\cellcolor[HTML]{FFF0A6}} \color[HTML]{000000} 59.02 ($\pm$0.91) \\
 &  & $(20, 10^{-2})$ & {\cellcolor[HTML]{07753E}} \color[HTML]{F1F1F1} 0.03 ($\pm$0.02) & {\cellcolor[HTML]{006837}} \color[HTML]{F1F1F1} \bfseries 0.00 & {\cellcolor[HTML]{DAF08D}} \color[HTML]{000000} 8.50 ($\pm$0.75) & {\cellcolor[HTML]{2AA054}} \color[HTML]{F1F1F1} 0.14 ($\pm$0.03) & {\cellcolor[HTML]{006837}} \color[HTML]{F1F1F1} \bfseries 0.68 ($\pm$0.05) & {\cellcolor[HTML]{F1F9AC}} \color[HTML]{000000} 60.77 ($\pm$0.41) \\
 &  & $(20, 10^{-3})$ & {\cellcolor[HTML]{ABDB6D}} \color[HTML]{000000} 0.31 ($\pm$0.11) & {\cellcolor[HTML]{0E8245}} \color[HTML]{F1F1F1} 5.47 & {\cellcolor[HTML]{FEDA86}} \color[HTML]{000000} 12.91 ($\pm$1.22) & {\cellcolor[HTML]{FDB567}} \color[HTML]{000000} 0.05 ($\pm$0.02) & {\cellcolor[HTML]{48AE5C}} \color[HTML]{F1F1F1} 0.57 ($\pm$0.06) & {\cellcolor[HTML]{F1F9AC}} \color[HTML]{000000} 60.73 ($\pm$0.25) \\
 \cmidrule{2-9}
 
 & \multirow[c]{2}{*}{GC} & $0.9$ & {\cellcolor[HTML]{F7FCB4}} \color[HTML]{000000} 0.48 ($\pm$0.08) & {\cellcolor[HTML]{F4FAB0}} \color[HTML]{000000} 46.88 & {\cellcolor[HTML]{FB9D59}} \color[HTML]{000000} 15.25 ($\pm$1.64) & {\cellcolor[HTML]{F16640}} \color[HTML]{F1F1F1} 0.03 ($\pm$0.01) & {\cellcolor[HTML]{ABDB6D}} \color[HTML]{000000} 0.47 ($\pm$0.07) & {\cellcolor[HTML]{F67A49}} \color[HTML]{F1F1F1} 54.38 ($\pm$1.73) \\
 &  & $0.99$ & {\cellcolor[HTML]{3FAA59}} \color[HTML]{F1F1F1} 0.15 ($\pm$0.04) & {\cellcolor[HTML]{006837}} \color[HTML]{F1F1F1} \bfseries 0.00 & {\cellcolor[HTML]{FEE999}} \color[HTML]{000000} 12.03 ($\pm$1.64) & {\cellcolor[HTML]{FEEC9F}} \color[HTML]{000000} 0.07 ($\pm$0.03) & {\cellcolor[HTML]{30A356}} \color[HTML]{F1F1F1} 0.59 ($\pm$0.05) & {\cellcolor[HTML]{A50026}} \color[HTML]{F1F1F1} \itshape 11.49 ($\pm$1.29) \\
 \cmidrule{2-9}
 
 & \multirow[c]{7}{*}{PRECODE} & $(1, 8)$ & {\cellcolor[HTML]{A5D86A}} \color[HTML]{000000} 0.30 ($\pm$0.09) & {\cellcolor[HTML]{016A38}} \color[HTML]{F1F1F1} 0.78 & {\cellcolor[HTML]{FBFDBA}} \color[HTML]{000000} 10.37 ($\pm$0.89) & {\cellcolor[HTML]{E6F59D}} \color[HTML]{000000} 0.09 ($\pm$0.02) & {\cellcolor[HTML]{3CA959}} \color[HTML]{F1F1F1} 0.58 ($\pm$0.07) & {\cellcolor[HTML]{FAFDB8}} \color[HTML]{000000} 60.24 ($\pm$0.90) \\
 &  & $(1, 16)$ & {\cellcolor[HTML]{A5D86A}} \color[HTML]{000000} 0.30 ($\pm$0.09) & {\cellcolor[HTML]{05713C}} \color[HTML]{F1F1F1} 2.34 & {\cellcolor[HTML]{FDFEBC}} \color[HTML]{000000} 10.40 ($\pm$0.83) & {\cellcolor[HTML]{E6F59D}} \color[HTML]{000000} 0.09 ($\pm$0.02) & {\cellcolor[HTML]{3CA959}} \color[HTML]{F1F1F1} 0.58 ($\pm$0.07) & {\cellcolor[HTML]{E9F6A1}} \color[HTML]{000000} 61.14 ($\pm$0.58) \\
 &  & $(1, 32)$ & {\cellcolor[HTML]{AFDD70}} \color[HTML]{000000} 0.32 ($\pm$0.09) & {\cellcolor[HTML]{036E3A}} \color[HTML]{F1F1F1} 1.56 & {\cellcolor[HTML]{FFFDBC}} \color[HTML]{000000} 10.67 ($\pm$0.80) & {\cellcolor[HTML]{E6F59D}} \color[HTML]{000000} 0.09 ($\pm$0.02) & {\cellcolor[HTML]{48AE5C}} \color[HTML]{F1F1F1} 0.57 ($\pm$0.07) & {\cellcolor[HTML]{E2F397}} \color[HTML]{000000} 61.49 ($\pm$0.51) \\
 &  & $(1, 64)$ & {\cellcolor[HTML]{BFE47A}} \color[HTML]{000000} 0.35 ($\pm$0.10) & {\cellcolor[HTML]{138C4A}} \color[HTML]{F1F1F1} 7.81 & {\cellcolor[HTML]{FFF3AC}} \color[HTML]{000000} 11.29 ($\pm$0.96) & {\cellcolor[HTML]{FEFFBE}} \color[HTML]{000000} 0.08 ($\pm$0.02) & {\cellcolor[HTML]{5DB961}} \color[HTML]{F1F1F1} 0.55 ($\pm$0.07) & {\cellcolor[HTML]{E0F295}} \color[HTML]{000000} 61.60 ($\pm$0.38) \\
 &  & $(1, 128)$ & {\cellcolor[HTML]{DCF08F}} \color[HTML]{000000} 0.41 ($\pm$0.11) & {\cellcolor[HTML]{60BA62}} \color[HTML]{F1F1F1} 19.53 & {\cellcolor[HTML]{FEE18D}} \color[HTML]{000000} 12.58 ($\pm$1.34) & {\cellcolor[HTML]{FED481}} \color[HTML]{000000} 0.06 ($\pm$0.02) & {\cellcolor[HTML]{8ECF67}} \color[HTML]{000000} 0.50 ($\pm$0.07) & {\cellcolor[HTML]{F2FAAE}} \color[HTML]{000000} 60.65 ($\pm$0.70) \\
 &  & $(1, 256)$ & {\cellcolor[HTML]{FFF6B0}} \color[HTML]{000000} 0.53 ($\pm$0.12) & {\cellcolor[HTML]{FED683}} \color[HTML]{000000} 61.72 & {\cellcolor[HTML]{FCA55D}} \color[HTML]{000000} 15.07 ($\pm$1.67) & {\cellcolor[HTML]{F16640}} \color[HTML]{F1F1F1} 0.03 ($\pm$0.02) & {\cellcolor[HTML]{CFEB85}} \color[HTML]{000000} 0.42 ($\pm$0.08) & {\cellcolor[HTML]{ECF7A6}} \color[HTML]{000000} 60.97 ($\pm$0.55) \\
 &  & $(1, 512)$ & {\cellcolor[HTML]{FBA05B}} \color[HTML]{000000} 0.72 ($\pm$0.09) & {\cellcolor[HTML]{AF0926}} \color[HTML]{F1F1F1} \itshape 97.66 & {\cellcolor[HTML]{D42D27}} \color[HTML]{F1F1F1} 19.07 ($\pm$1.78) & {\cellcolor[HTML]{C41E27}} \color[HTML]{F1F1F1} \itshape 0.01 ($\pm$0.01) & {\cellcolor[HTML]{FEDE89}} \color[HTML]{000000} 0.27 ($\pm$0.08) & {\cellcolor[HTML]{FFFEBE}} \color[HTML]{000000} 59.95 ($\pm$0.11) \\
\cmidrule{2-9}

 & \textbf{Ours} & $(1, 5,\frac{1}{2}, 10^{-1})$ & {\cellcolor[HTML]{6BBF64}} \color[HTML]{000000} 0.21 ($\pm$0.08) & {\cellcolor[HTML]{006837}} \color[HTML]{F1F1F1} \bfseries 0.00 & {\cellcolor[HTML]{ECF7A6}} \color[HTML]{000000} 9.51 ($\pm$1.49) & {\cellcolor[HTML]{84CA66}} \color[HTML]{000000} 0.12 ($\pm$0.04) & {\cellcolor[HTML]{3CA959}} \color[HTML]{F1F1F1} 0.58 ($\pm$0.06) & {\cellcolor[HTML]{39A758}} \color[HTML]{F1F1F1} \bfseries 67.17 ($\pm$0.58) \\

\bottomrule
\end{tabular}
\end{table*}

\begin{table*}
\centering
\caption{
Privacy and model utility metrics for a CNN, ResNet-18 and ViT without and with our proposed CVB defense mechanism trained on MNIST and CIFAR-10. 
When our CVB defense is used, our proposed targeted attack is applied.
Arrows indicate direction of improvement. 
Bold and italic formatting highlight best and worst results respectively.
}
\label{tab:apx_model}
\resizebox{0.99\linewidth}{!}{%
\begin{tabular}{c|c|c|c|c|c|c|c|c|c|c|c|c}
\toprule
 & Model & Defense & $P$ & $s_E$ & $k_E$ & $\beta$ & SSIM $\downarrow$ & ASR [\%] $\downarrow$ & PSNR $\downarrow$ & MSE $\uparrow$ & LPIPS $\uparrow$ & Accuracy [\%] $\uparrow$ \\
\midrule
\multirow[c]{8}{*}{\rotatebox{90}{MNIST}} & \multirow[c]{2}{*}{CNN} & None & - & - & - & - & {\cellcolor[HTML]{BD1726}} \color[HTML]{F1F1F1} 0.95 ($\pm$0.06) & {\cellcolor[HTML]{A50026}} \color[HTML]{F1F1F1} \itshape 100.00 & {\cellcolor[HTML]{A50026}} \color[HTML]{F1F1F1} \itshape 33.13 ($\pm$7.63) & {\cellcolor[HTML]{A50026}} \color[HTML]{F1F1F1} \itshape 0.00 ($\pm$0.00) & {\cellcolor[HTML]{B91326}} \color[HTML]{F1F1F1} 0.03 ($\pm$0.06) & {\cellcolor[HTML]{006837}} \color[HTML]{F1F1F1} 99.10 ($\pm$0.04) \\
 &   & Ours &  1 & $\frac{1}{2}$  & 5 & $10^{-1}$ & {\cellcolor[HTML]{A0D669}} \color[HTML]{000000} 0.29 ($\pm$0.09) & {\cellcolor[HTML]{016A38}} \color[HTML]{F1F1F1} 0.78 & {\cellcolor[HTML]{73C264}} \color[HTML]{000000} \bfseries 7.34 ($\pm$1.55) & {\cellcolor[HTML]{006837}} \color[HTML]{F1F1F1} \bfseries 0.20 ($\pm$0.07) & {\cellcolor[HTML]{199750}} \color[HTML]{F1F1F1} 0.64 ($\pm$0.05) & {\cellcolor[HTML]{006837}} \color[HTML]{F1F1F1} 99.15 ($\pm$0.05) \\
\cmidrule{2-13}
 & \multirow[c]{2}{*}{ResNet-18} & None & - & - & - & - & {\cellcolor[HTML]{F4FAB0}} \color[HTML]{000000} 0.47 ($\pm$0.11) & {\cellcolor[HTML]{CDEA83}} \color[HTML]{000000} 37.50 & {\cellcolor[HTML]{D9EF8B}} \color[HTML]{000000} 13.30 ($\pm$2.31) & {\cellcolor[HTML]{F98E52}} \color[HTML]{F1F1F1} 0.05 ($\pm$0.03) & {\cellcolor[HTML]{FDFEBC}} \color[HTML]{000000} 0.36 ($\pm$0.12) & {\cellcolor[HTML]{006837}} \color[HTML]{F1F1F1} 99.39 ($\pm$0.09) \\
 &   & Ours &  1 & $\frac{1}{2}$  & 5 & $5\cdot10^{-5}$ & {\cellcolor[HTML]{0A7B41}} \color[HTML]{F1F1F1} \bfseries 0.04 ($\pm$0.05) & {\cellcolor[HTML]{006837}} \color[HTML]{F1F1F1} \bfseries 0.00 & {\cellcolor[HTML]{98D368}} \color[HTML]{000000} 9.19 ($\pm$1.55) & {\cellcolor[HTML]{BFE47A}} \color[HTML]{000000} 0.13 ($\pm$0.05) & {\cellcolor[HTML]{128A49}} \color[HTML]{F1F1F1} 0.66 ($\pm$0.07) & {\cellcolor[HTML]{006837}} \color[HTML]{F1F1F1} \bfseries 99.46 ($\pm$0.04) \\
\cmidrule{2-13}
 & \multirow[c]{2}{*}{ViT} & None & - & - & - & - & {\cellcolor[HTML]{A90426}} \color[HTML]{F1F1F1} \itshape 0.99 ($\pm$0.00) & {\cellcolor[HTML]{A50026}} \color[HTML]{F1F1F1} \itshape 100.00 & {\cellcolor[HTML]{BD1726}} \color[HTML]{F1F1F1} 31.56 ($\pm$2.51) & {\cellcolor[HTML]{A50026}} \color[HTML]{F1F1F1} \itshape 0.00 ($\pm$0.00) & {\cellcolor[HTML]{AB0626}} \color[HTML]{F1F1F1} \itshape 0.01 ($\pm$0.01) & {\cellcolor[HTML]{006837}} \color[HTML]{F1F1F1} \itshape 98.70 ($\pm$0.14) \\
 &   & Ours &  1 & $\frac{1}{2}$  & 3 & $10^{-1}$ & {\cellcolor[HTML]{C9E881}} \color[HTML]{000000} 0.37 ($\pm$0.08) & {\cellcolor[HTML]{118848}} \color[HTML]{F1F1F1} 7.03 & {\cellcolor[HTML]{C5E67E}} \color[HTML]{000000} 11.97 ($\pm$0.54) & {\cellcolor[HTML]{FDAD60}} \color[HTML]{000000} 0.06 ($\pm$0.01) & {\cellcolor[HTML]{006837}} \color[HTML]{F1F1F1} \bfseries 0.71 ($\pm$0.04) & {\cellcolor[HTML]{006837}} \color[HTML]{F1F1F1} 98.96 ($\pm$0.04) \\

\midrule

\multirow[c]{8}{*}{\rotatebox{90}{CIFAR-10}} & \multirow[c]{2}{*}{CNN} & None & - & - & - & - & {\cellcolor[HTML]{E0422F}} \color[HTML]{F1F1F1} 0.87 ($\pm$0.11) & {\cellcolor[HTML]{B30D26}} \color[HTML]{F1F1F1} 96.88 & {\cellcolor[HTML]{F16640}} \color[HTML]{F1F1F1} 21.09 ($\pm$2.88) & {\cellcolor[HTML]{CE2827}} \color[HTML]{F1F1F1} 0.01 ($\pm$0.01) & {\cellcolor[HTML]{FDB768}} \color[HTML]{000000} 0.20 ($\pm$0.09) & {\cellcolor[HTML]{CBE982}} \color[HTML]{000000} \itshape 62.53 ($\pm$0.16) \\
 &   & Ours &  1 & $\frac{1}{2}$  & 5 & $10^{-1}$ & {\cellcolor[HTML]{6BBF64}} \color[HTML]{000000} 0.21 ($\pm$0.08) & {\cellcolor[HTML]{006837}} \color[HTML]{F1F1F1} \bfseries 0.00 & {\cellcolor[HTML]{C7E77F}} \color[HTML]{000000} \bfseries 9.51 ($\pm$1.49) & {\cellcolor[HTML]{006837}} \color[HTML]{F1F1F1} \bfseries 0.12 ($\pm$0.04) & {\cellcolor[HTML]{148E4B}} \color[HTML]{F1F1F1} 0.58 ($\pm$0.06) & {\cellcolor[HTML]{39A758}} \color[HTML]{F1F1F1} 67.17 ($\pm$0.58) \\
\cmidrule{2-13}
 & \multirow[c]{2}{*}{ResNet-18} & None & - & - & - & - & {\cellcolor[HTML]{FFF0A6}} \color[HTML]{000000} 0.55 ($\pm$0.11) & {\cellcolor[HTML]{FDB365}} \color[HTML]{000000} 68.75 & {\cellcolor[HTML]{FDC776}} \color[HTML]{000000} 16.89 ($\pm$2.39) & {\cellcolor[HTML]{EA5739}} \color[HTML]{F1F1F1} 0.02 ($\pm$0.01) & {\cellcolor[HTML]{E3F399}} \color[HTML]{000000} 0.36 ($\pm$0.08) & {\cellcolor[HTML]{006837}} \color[HTML]{F1F1F1} 73.18 ($\pm$0.37) \\
 &   & Ours &  1 & $\frac{1}{2}$  & 5 & $5\cdot10^{-5}$ & {\cellcolor[HTML]{17934E}} \color[HTML]{F1F1F1} \bfseries 0.09 ($\pm$0.07) & {\cellcolor[HTML]{006837}} \color[HTML]{F1F1F1} \bfseries 0.00 & {\cellcolor[HTML]{E6F59D}} \color[HTML]{000000} 11.30 ($\pm$2.04) & {\cellcolor[HTML]{B7E075}} \color[HTML]{000000} 0.08 ($\pm$0.04) & {\cellcolor[HTML]{108647}} \color[HTML]{F1F1F1} 0.59 ($\pm$0.06) & {\cellcolor[HTML]{006837}} \color[HTML]{F1F1F1} \bfseries 74.17 ($\pm$0.20) \\
\cmidrule{2-13}
 & \multirow[c]{2}{*}{ViT} & None & - & - & - & - & {\cellcolor[HTML]{D62F27}} \color[HTML]{F1F1F1} \itshape 0.90 ($\pm$0.05) & {\cellcolor[HTML]{A50026}} \color[HTML]{F1F1F1} \itshape 100.00 & {\cellcolor[HTML]{A50026}} \color[HTML]{F1F1F1} \itshape 25.99 ($\pm$1.80) & {\cellcolor[HTML]{A50026}} \color[HTML]{F1F1F1} \itshape 0.00 ($\pm$0.00) & {\cellcolor[HTML]{EC5C3B}} \color[HTML]{F1F1F1} \itshape 0.11 ($\pm$0.05) & {\cellcolor[HTML]{B5DF74}} \color[HTML]{000000} 63.39 ($\pm$0.50) \\
 &   & Ours &  1 & $\frac{1}{2}$  & 3 & $10^{-1}$ & {\cellcolor[HTML]{45AD5B}} \color[HTML]{F1F1F1} 0.16 ($\pm$0.06) & {\cellcolor[HTML]{006837}} \color[HTML]{F1F1F1} \bfseries 0.00 & {\cellcolor[HTML]{C9E881}} \color[HTML]{000000} 9.63 ($\pm$0.54) & {\cellcolor[HTML]{15904C}} \color[HTML]{F1F1F1} 0.11 ($\pm$0.02) & {\cellcolor[HTML]{006837}} \color[HTML]{F1F1F1} \bfseries 0.63 ($\pm$0.06) & {\cellcolor[HTML]{B3DF72}} \color[HTML]{000000} 63.50 ($\pm$0.34) \\
\bottomrule
\end{tabular}
}
\end{table*}

\begin{table*}
\centering
\caption{
Privacy and model utility metrics for a CNN without and with our proposed CVB defense mechanism trained on four MedMNIST datasets. 
When our CVB defense is used, our proposed targeted attack is applied.
Arrows indicate direction of improvement. 
Bold and italic formatting highlight best and worst results respectively.
}
\label{tab:apx_datasets}
\resizebox{0.99\linewidth}{!}{%

\begin{tabular}{c|c|c|c|c|c|c|c|c|c|c|c}
\toprule
 Datset & Defense & $P$ & $s_E$ & $k_E$ & $\beta$ & SSIM $\downarrow$ & ASR [\%] $\downarrow$ & PSNR $\downarrow$ & MSE $\uparrow$ & LPIPS $\uparrow$ & Accuracy [\%] $\uparrow$ \\
\midrule
\multirow[c]{2}{*}{Pneumonia} & None & - & - & - & - & {\cellcolor[HTML]{BD1726}} \color[HTML]{F1F1F1} \itshape 0.95 ($\pm$0.04) & {\cellcolor[HTML]{A50026}} \color[HTML]{F1F1F1} \itshape 100.00 & {\cellcolor[HTML]{A50026}} \color[HTML]{F1F1F1} \itshape 21.24 ($\pm$2.91) & {\cellcolor[HTML]{E34933}} \color[HTML]{F1F1F1} \itshape 0.01 ($\pm$0.01) & {\cellcolor[HTML]{FCA55D}} \color[HTML]{000000} \itshape 0.15 ($\pm$0.05) & {\cellcolor[HTML]{006837}} \color[HTML]{F1F1F1} \itshape 86.92 ($\pm$0.21) \\
 & Ours & 1 & $\frac{1}{2}$ & 5 & $10^{-3}$ & {\cellcolor[HTML]{BFE47A}} \color[HTML]{000000} \bfseries 0.35 ($\pm$0.11) & {\cellcolor[HTML]{138C4A}} \color[HTML]{F1F1F1} \bfseries 7.81 & {\cellcolor[HTML]{FEEB9D}} \color[HTML]{000000} \bfseries 11.97 ($\pm$1.74) & {\cellcolor[HTML]{006837}} \color[HTML]{F1F1F1} \bfseries 0.07 ($\pm$0.03) & {\cellcolor[HTML]{006837}} \color[HTML]{F1F1F1} \bfseries 0.52 ($\pm$0.06) & {\cellcolor[HTML]{006837}} \color[HTML]{F1F1F1} \bfseries 87.31 ($\pm$1.22) \\
\midrule\multirow[c]{2}{*}{Derma} & None & - & - & - & - & {\cellcolor[HTML]{E24731}} \color[HTML]{F1F1F1} \itshape 0.86 ($\pm$0.13) & {\cellcolor[HTML]{B30D26}} \color[HTML]{F1F1F1} \itshape 96.88 & {\cellcolor[HTML]{A50026}} \color[HTML]{F1F1F1} \itshape 24.77 ($\pm$4.88) & {\cellcolor[HTML]{E34933}} \color[HTML]{F1F1F1} \itshape 0.01 ($\pm$0.01) & {\cellcolor[HTML]{FB9D59}} \color[HTML]{000000} \itshape 0.16 ($\pm$0.11) & {\cellcolor[HTML]{006837}} \color[HTML]{F1F1F1} \bfseries 75.02 ($\pm$0.40) \\
 & Ours & 1 & $\frac{1}{2}$ & 3 & $10^{-1}$ & {\cellcolor[HTML]{5DB961}} \color[HTML]{F1F1F1} \bfseries 0.19 ($\pm$0.10) & {\cellcolor[HTML]{006837}} \color[HTML]{F1F1F1} \bfseries 0.00 & {\cellcolor[HTML]{F5FBB2}} \color[HTML]{000000} \bfseries 11.71 ($\pm$1.88) & {\cellcolor[HTML]{006837}} \color[HTML]{F1F1F1} \bfseries 0.07 ($\pm$0.03) & {\cellcolor[HTML]{006837}} \color[HTML]{F1F1F1} \bfseries 0.58 ($\pm$0.06) & {\cellcolor[HTML]{006837}} \color[HTML]{F1F1F1} \itshape 74.87 ($\pm$0.69) \\

\midrule\multirow[c]{2}{*}{Retina} & None & - & - & - & - & {\cellcolor[HTML]{E54E35}} \color[HTML]{F1F1F1} \itshape 0.85 ($\pm$0.14) & {\cellcolor[HTML]{B91326}} \color[HTML]{F1F1F1} \itshape 96.09 & {\cellcolor[HTML]{A50026}} \color[HTML]{F1F1F1} \itshape 18.40 ($\pm$2.56) & {\cellcolor[HTML]{F67A49}} \color[HTML]{F1F1F1} \itshape 0.02 ($\pm$0.01) & {\cellcolor[HTML]{FECE7C}} \color[HTML]{000000} \itshape 0.24 ($\pm$0.10) & {\cellcolor[HTML]{006837}} \color[HTML]{F1F1F1} \itshape 54.62 ($\pm$1.97) \\
 & Ours & 1 & $\frac{1}{2}$ & 3 & $10^{-3}$ & {\cellcolor[HTML]{30A356}} \color[HTML]{F1F1F1} \bfseries 0.13 ($\pm$0.06) & {\cellcolor[HTML]{006837}} \color[HTML]{F1F1F1} \bfseries 0.00 & {\cellcolor[HTML]{FEE18D}} \color[HTML]{000000} \bfseries 10.99 ($\pm$1.92) & {\cellcolor[HTML]{006837}} \color[HTML]{F1F1F1} \bfseries 0.09 ($\pm$0.04) & {\cellcolor[HTML]{006837}} \color[HTML]{F1F1F1} \bfseries 0.66 ($\pm$0.04) & {\cellcolor[HTML]{006837}} \color[HTML]{F1F1F1} \bfseries 54.99 ($\pm$1.00) \\

\midrule\multirow[c]{2}{*}{Blood} & None & - & - & - & - & {\cellcolor[HTML]{E0422F}} \color[HTML]{F1F1F1} \itshape 0.87 ($\pm$0.11) & {\cellcolor[HTML]{B30D26}} \color[HTML]{F1F1F1} \itshape 96.88 & {\cellcolor[HTML]{A50026}} \color[HTML]{F1F1F1} \itshape 22.90 ($\pm$3.06) & {\cellcolor[HTML]{DE402E}} \color[HTML]{F1F1F1} \itshape 0.01 ($\pm$0.01) & {\cellcolor[HTML]{FB9D59}} \color[HTML]{000000} \itshape 0.16 ($\pm$0.09) & {\cellcolor[HTML]{006837}} \color[HTML]{F1F1F1} \itshape 89.82 ($\pm$0.08) \\
 & Ours & 1 & $\frac{1}{2}$ & 5 & $10^{-1}$ & {\cellcolor[HTML]{78C565}} \color[HTML]{000000} \bfseries 0.23 ($\pm$0.08) & {\cellcolor[HTML]{006837}} \color[HTML]{F1F1F1} \bfseries 0.00 & {\cellcolor[HTML]{FAFDB8}} \color[HTML]{000000} \bfseries 11.10 ($\pm$1.16) & {\cellcolor[HTML]{006837}} \color[HTML]{F1F1F1} \bfseries 0.08 ($\pm$0.02) & {\cellcolor[HTML]{006837}} \color[HTML]{F1F1F1} \bfseries 0.58 ($\pm$0.04) & {\cellcolor[HTML]{006837}} \color[HTML]{F1F1F1} \bfseries 92.34 ($\pm$0.13) \\
\bottomrule
\end{tabular}
}
\end{table*}

\end{document}